\documentclass[10pt,twocolumn,letterpaper]{article}

\usepackage{wacv}
\usepackage{times}
\usepackage{epsfig}
\usepackage{graphicx}
\usepackage{amsmath}
\usepackage{amssymb}
\usepackage{enumerate}
\usepackage{multirow}
\usepackage{color}
\usepackage{balance}

\usepackage[pagebackref=true,breaklinks=true,letterpaper=true,colorlinks,bookmarks=false]{hyperref}

\wacvfinalcopy 

\def\wacvPaperID{112} 

\begin{document}

\title{\large Depth Map Completion by Jointly Exploiting Blurry Color Images and Sparse Depth Maps}

\author{Liyuan Pan $^{1,2}$, Yuchao Dai$^{1,2}$, Miaomiao Liu$^{3,2}$ and Fatih Porikli$^{2}$ \\
$^{1}$ Northwestern Polytechnical University, Xi'an, China \\
$^{2}$ Australian National University, Canberra, Australia \\
$^{3}$ Data61, CSIRO, Canberra, Australia \\\tt\small{panliyuan@mail.nwpu.edu.cn, \{yuchao.dai, miaomiao.liu, fatih.porikli\}}@anu.edu.au}
\maketitle

\begin{abstract}
We aim at predicting a complete and high-resolution depth map from incomplete, sparse and noisy depth measurements. Existing methods handle this problem either by exploiting various regularizations on the depth maps directly or resorting to learning based methods. When the corresponding color images are available, the correlation between the depth maps and the color images are used to improve the completion performance, assuming the color images are clean and sharp. However, in real world dynamic scenes, color images are often blurry due to the camera motion and the moving objects in the scene. In this paper, we propose to tackle the problem of depth map completion by jointly exploiting the blurry color image sequences and the sparse depth map measurements, and present an energy minimization based formulation to simultaneously complete the depth maps, estimate the scene flow and deblur the color images. Our experimental evaluations on both outdoor and indoor scenarios demonstrate the state-of-the-art performance of our approach.
\end{abstract}

\section{Introduction}
\begin{figure}
\begin{center}
\begin{tabular}{cc}
\hspace{0.0cm}
\includegraphics[width=0.19\textwidth]{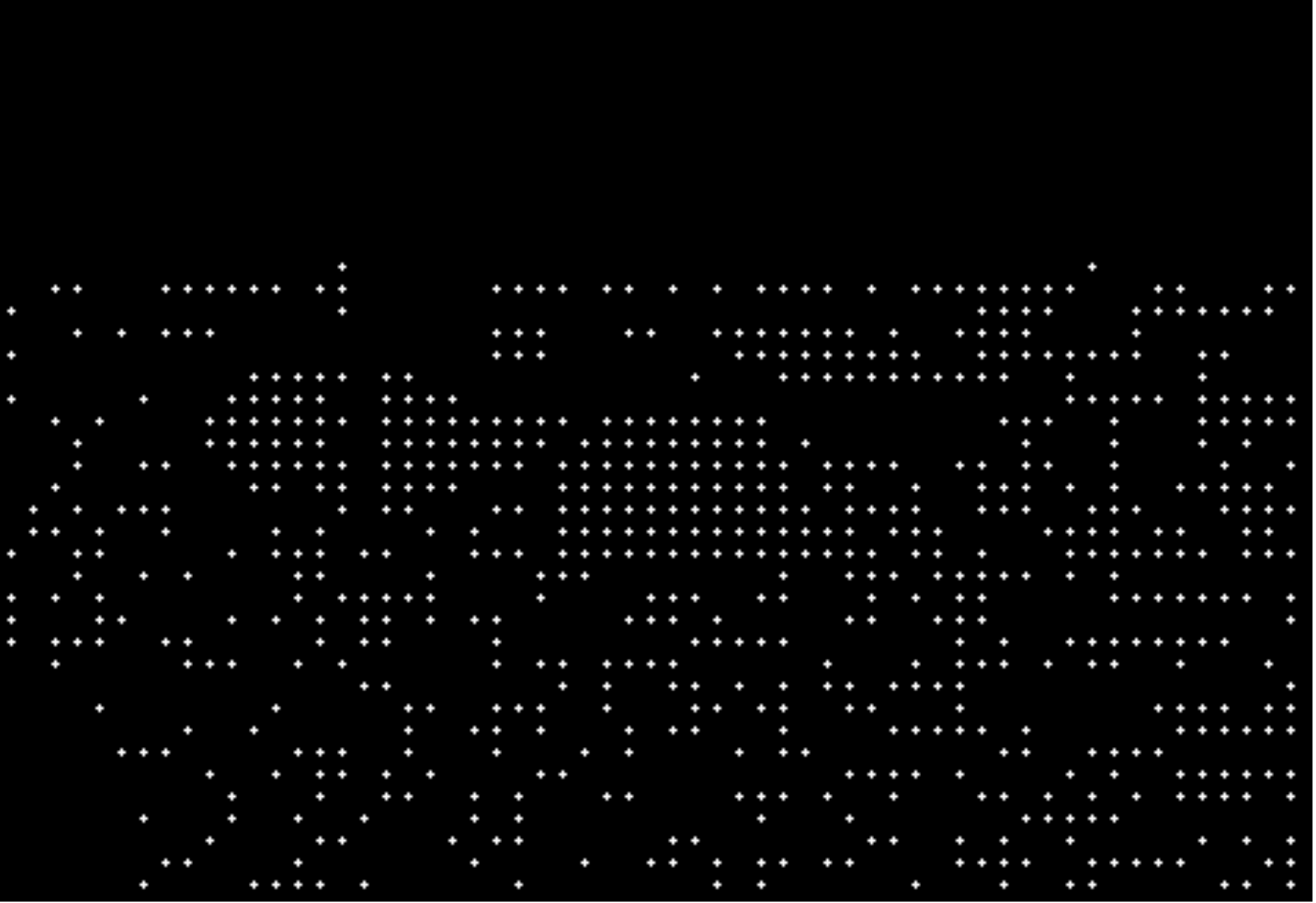}
&\hspace{0.0cm}
\includegraphics[width=0.19\textwidth]{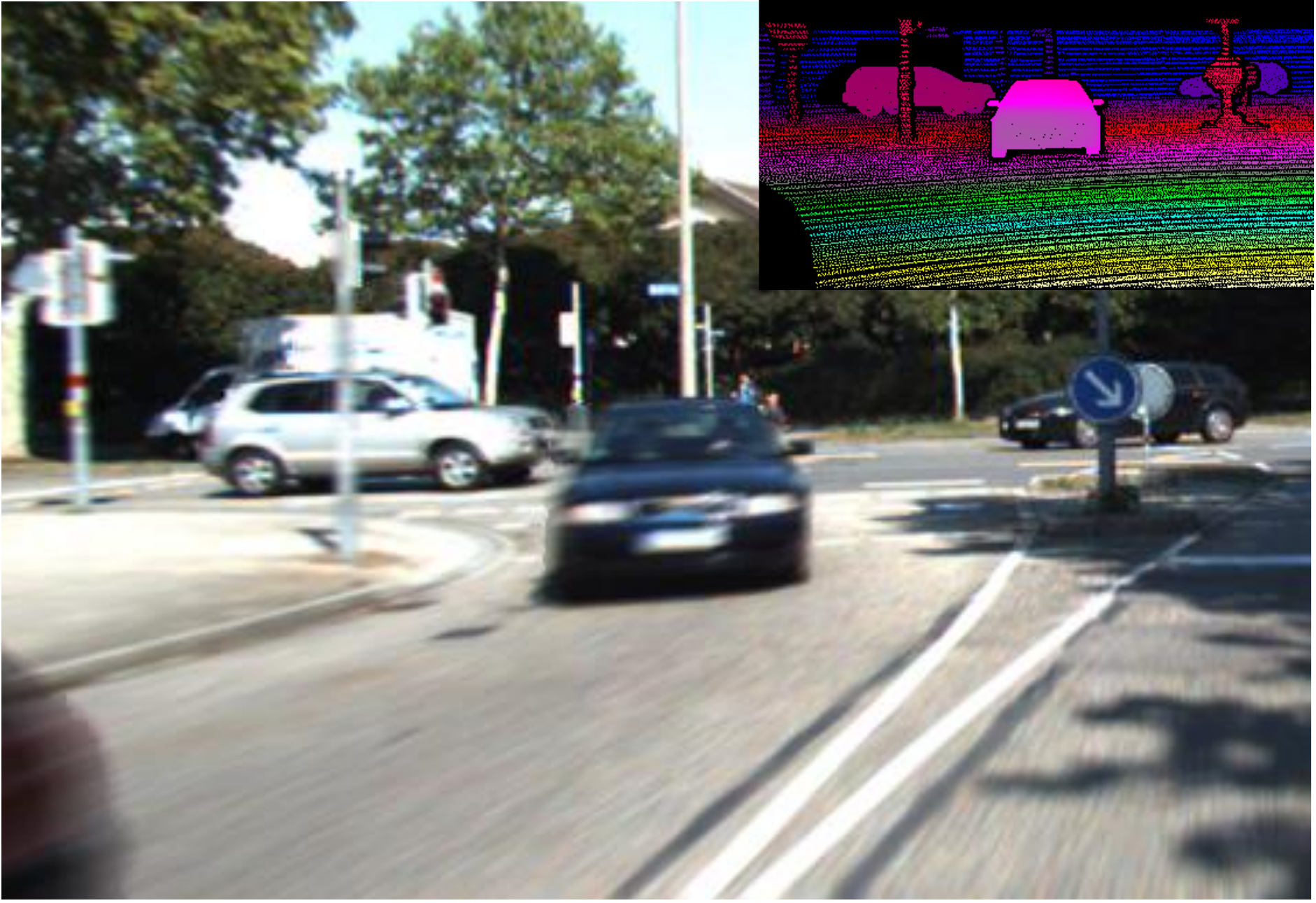}\\
(a) Input sparse depth map  &(b) Input blurry image\\
\hspace{0.0cm}
\includegraphics[width=0.19\textwidth]{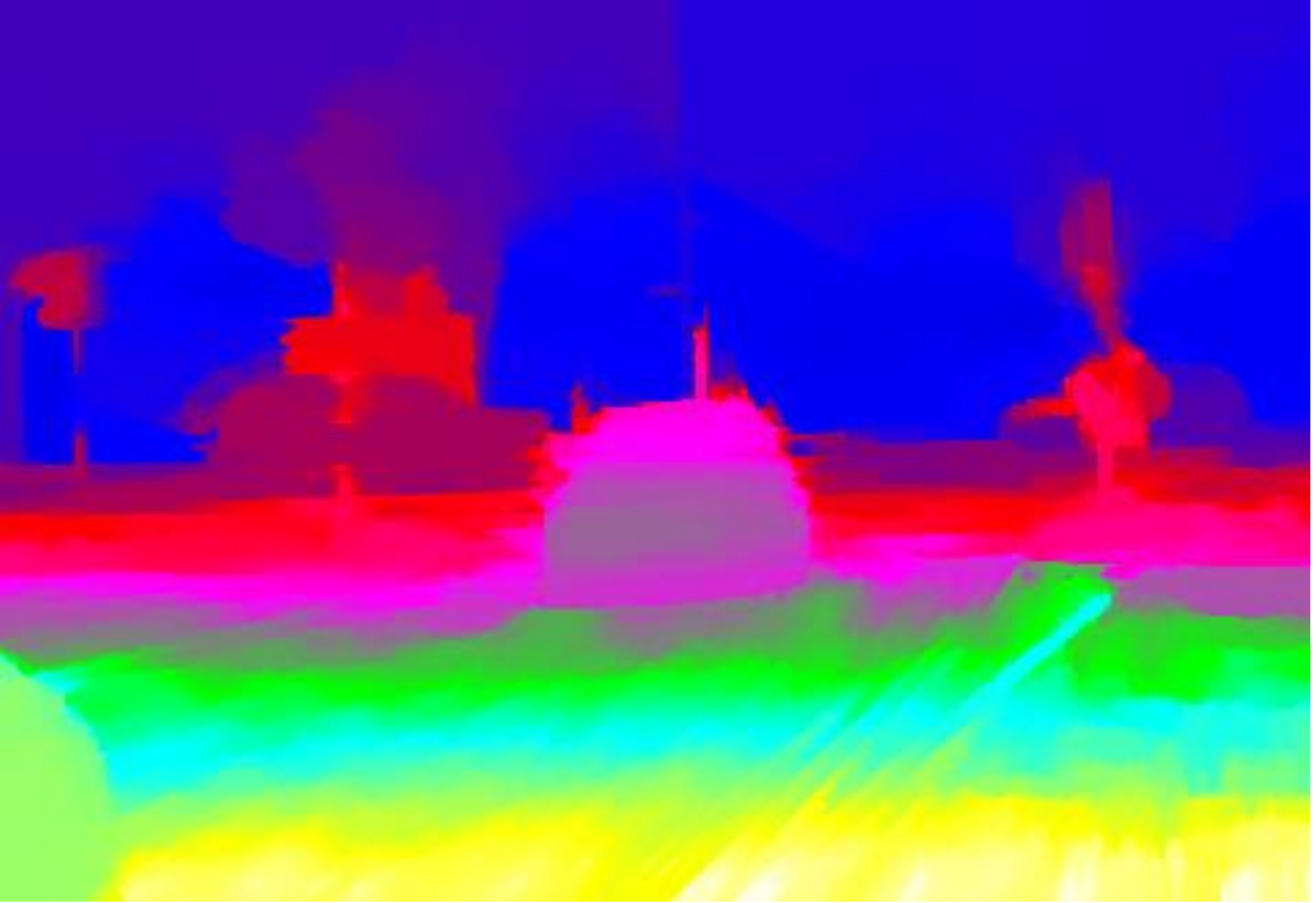}
&\hspace{0.0cm}
\includegraphics[width=0.19\textwidth]{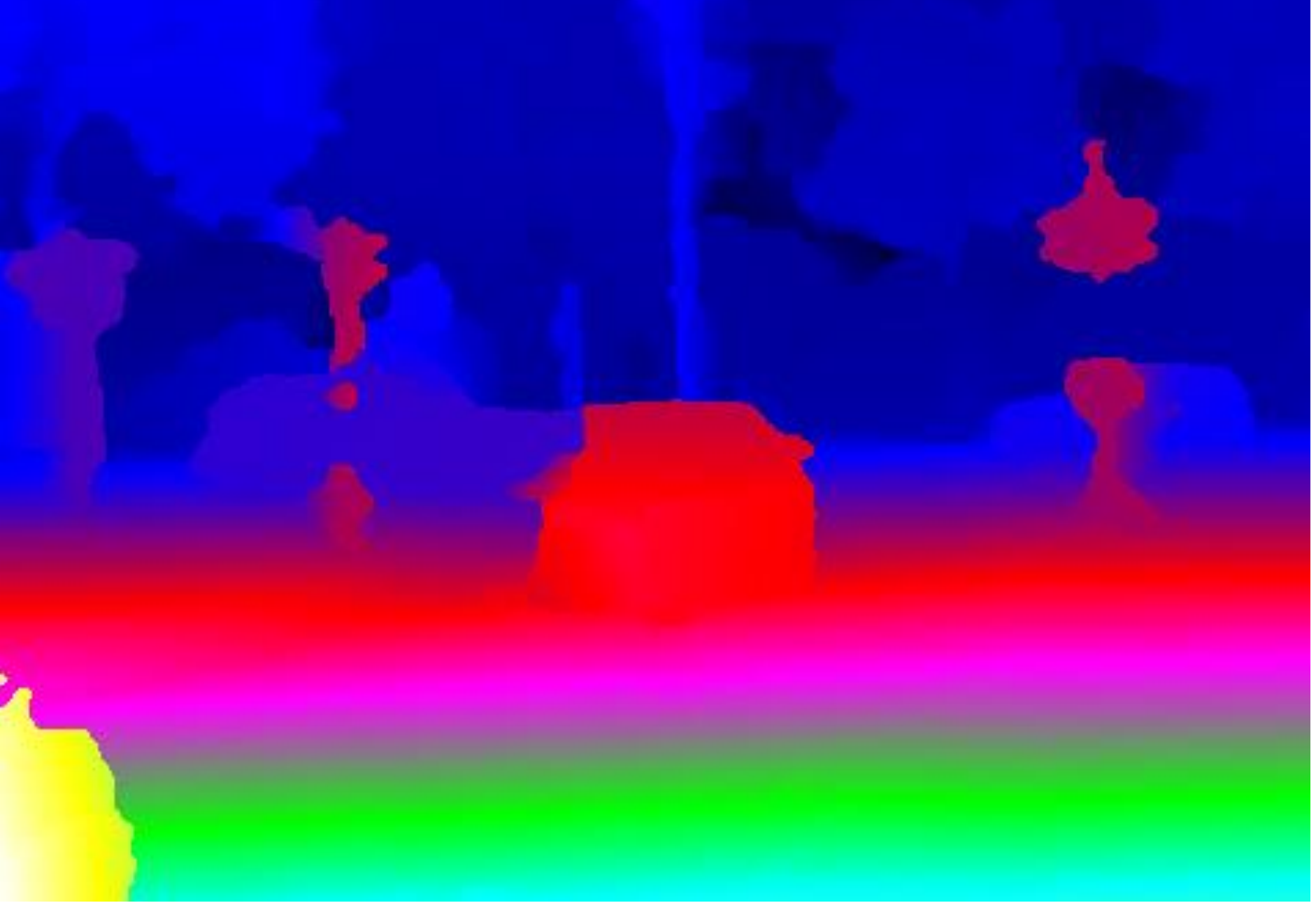}\\
(c) Park \etal \cite{park2014high} &(d) Vogel \etal \cite{vogel20153d}\\
\hspace{0.0cm}
\includegraphics[width=0.19\textwidth]{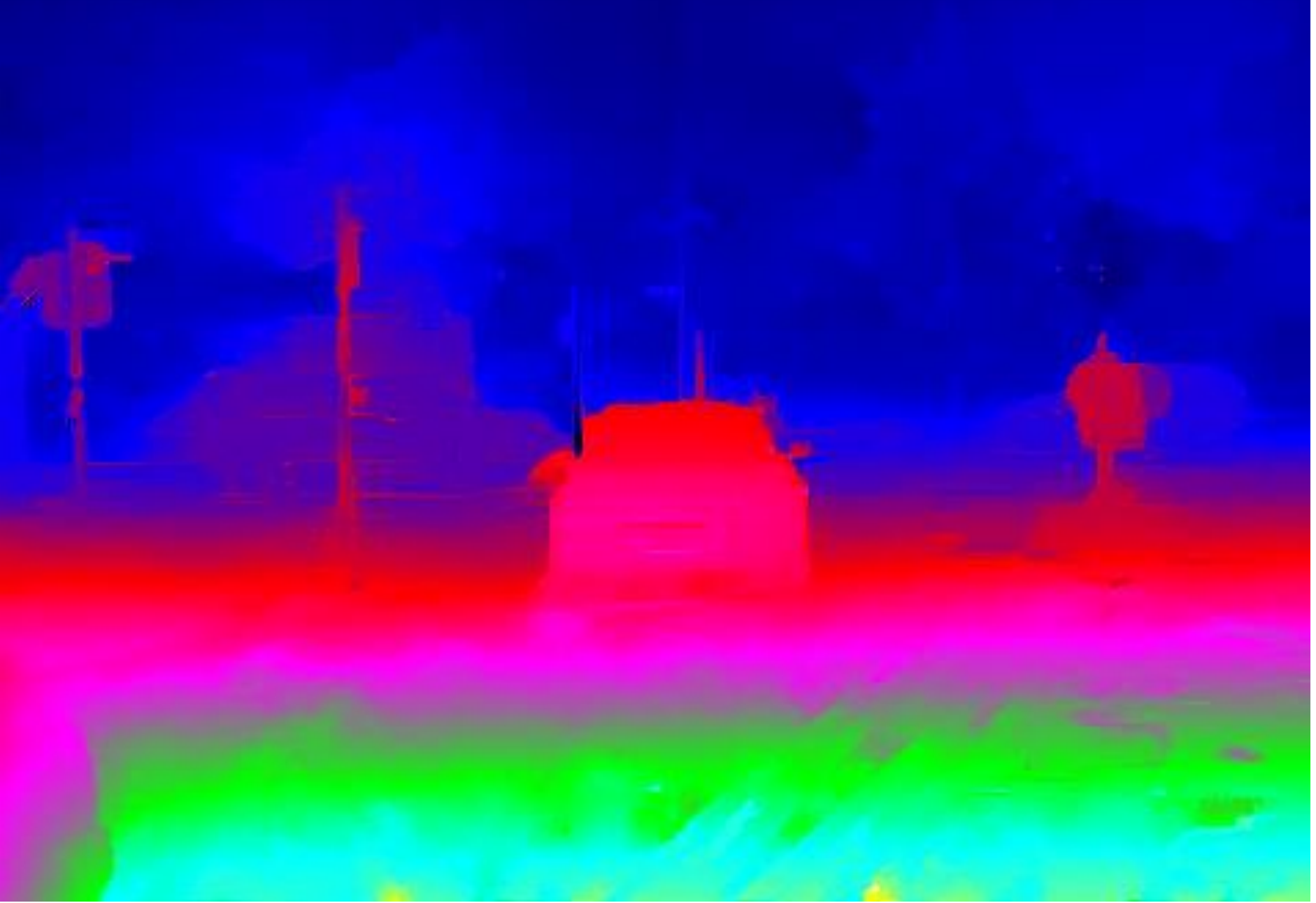}
&\hspace{0.0cm}
\includegraphics[width=0.19\textwidth]{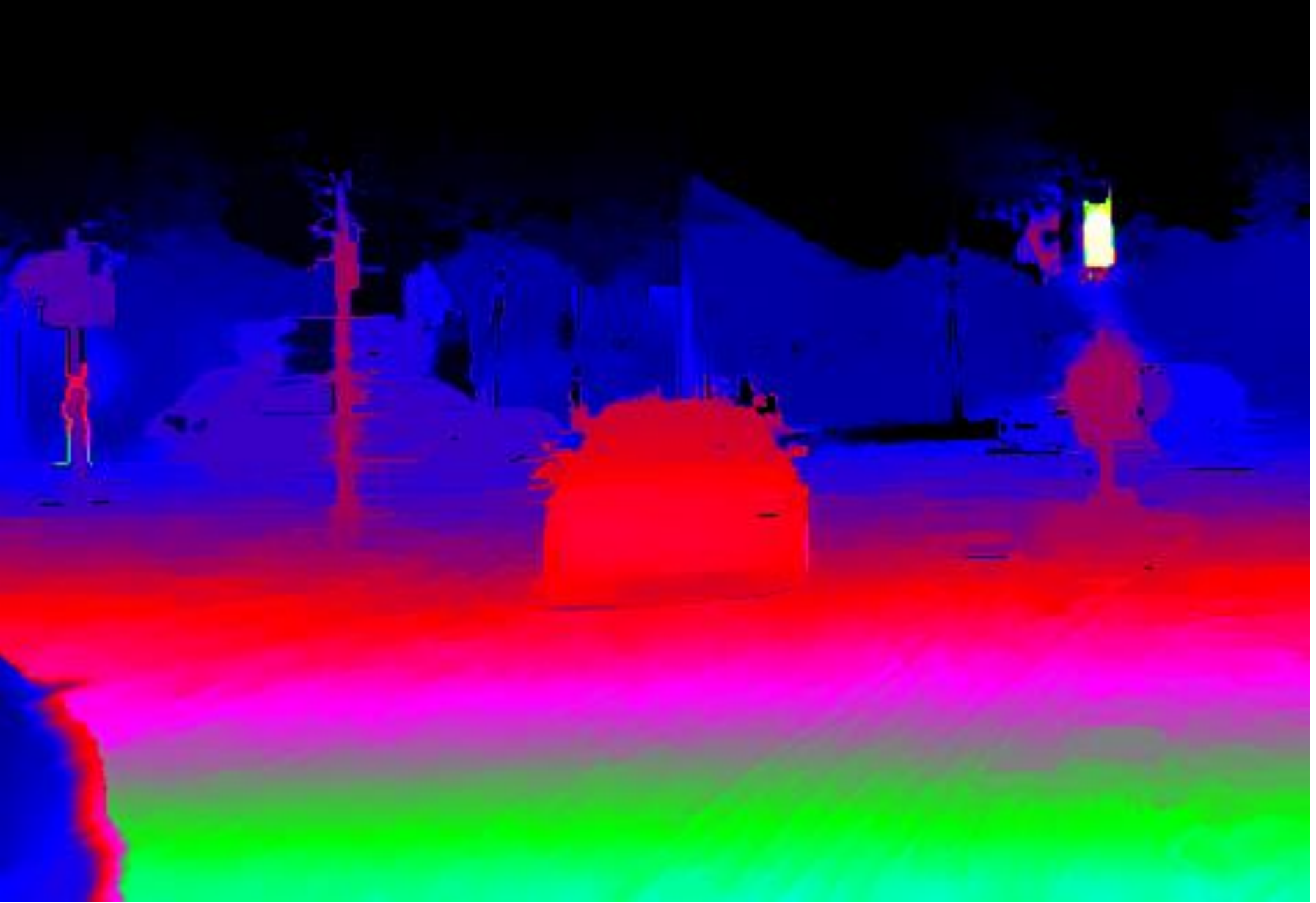}\\
(e) Ferstl \etal \cite{ferstl2013image}   &(f) Yang \etal ~\cite{yang2014color}\\
\hspace{0.0cm}
\includegraphics[width=0.19\textwidth]{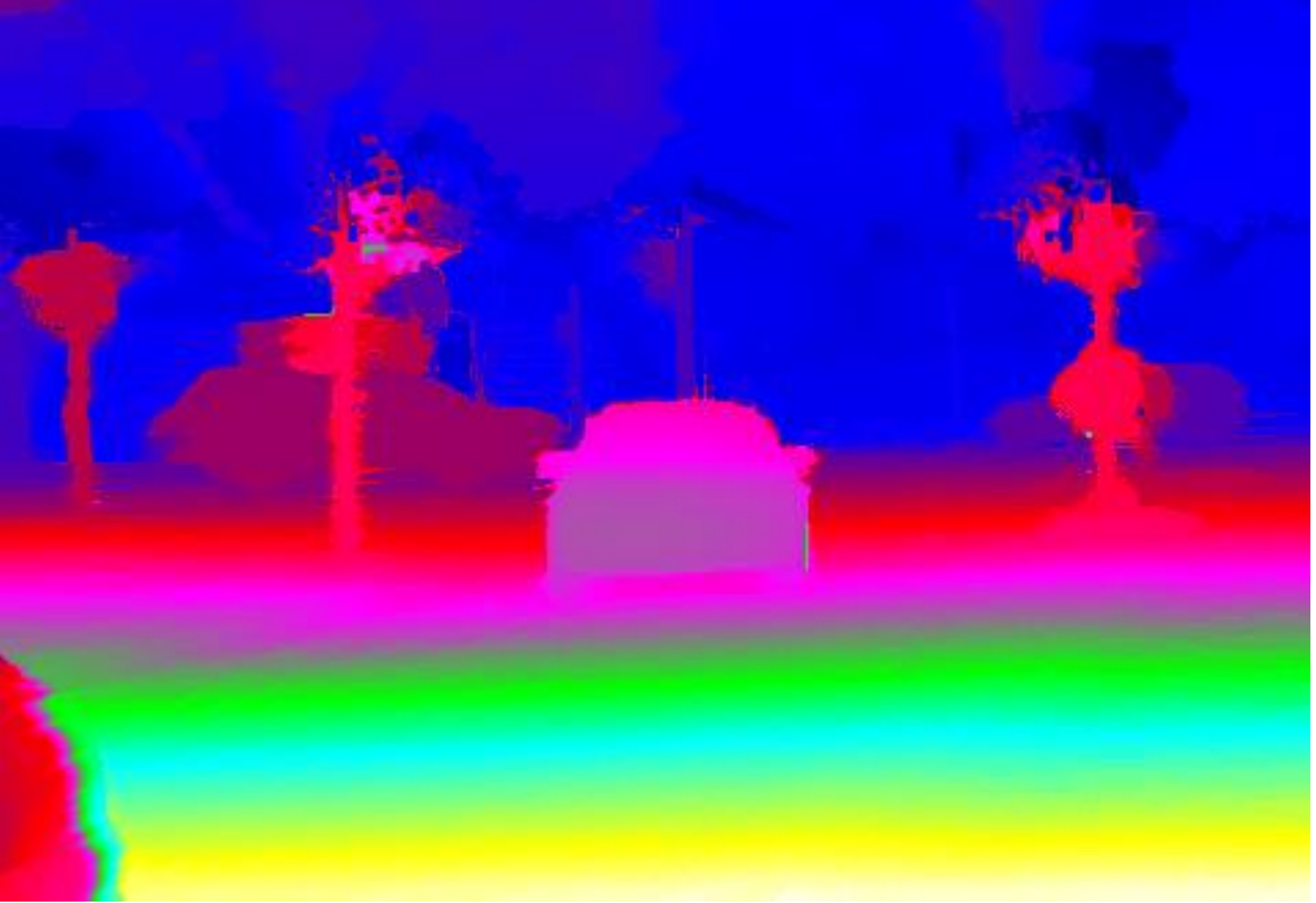}
&\hspace{0.0cm}
\includegraphics[width=0.19\textwidth]{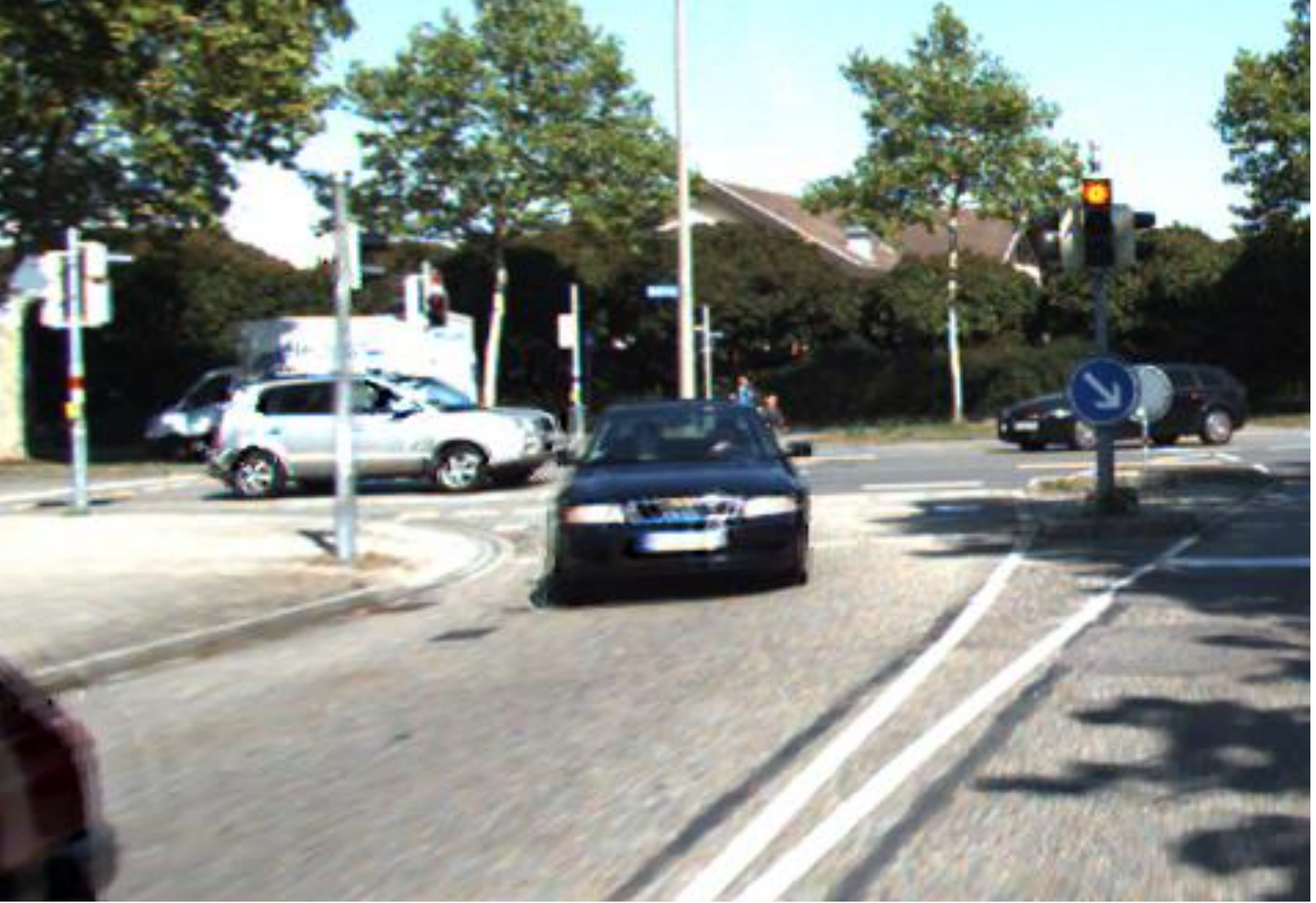}\\
\multicolumn{2}{c}{(g) Ours: completed depth map and deblurred image}  \\
\end{tabular}
\end{center}
\caption{Qualitative comparisons on depth completion performance. (a) Input: incomplete and noisy depth map. (b) Corresponding blurry color image (with ground-truth complete depth map overlaid in the corner). (c) Estimated depth by \cite{park2014high} (d) Estimated depth by \cite{vogel20153d}. (e) Estimated depth by \cite{ferstl2013image}. (f) Estimated depth by \cite{yang2014color}. (g) Our depth completion and deblurring result. Compared to the stereo method (i.e. \cite{vogel20153d}) that ranks as the $1^{st}$ on the KITTI dataset and the remaining three state-of-the-art depth completion methods (i.e. \cite{park2014high,ferstl2013image, yang2014color}) shown above, our method achieves the best performance. (Best viewed on screen). }
\label{fig:first}
\vspace{-5mm}
\end{figure}

High-precision and high-resolution 3D information play significant role in a variety of computer vision tasks including autonomous navigation~\cite{geiger2012we,Liu_2017_ITS}, 3D reconstruction and modeling~\cite{Liu_2017_ICCV, sun2010depth}, and image deblurring~\cite{arun2015multi, hu2014joint, seok2013dense, xu2012depth} just to count a few. However, the acquisition of such accurate depth maps is a challenging task. Although high-resolution depth maps can be computed from stereo images, the quality of the depth map relies on the calibration process and the apparent scene flow. Besides, stereoscopic depth estimation is problematic in low texture areas. As an alternative, active depth sensors provide depth information in a single shot. Unfortunately, measurements from the best depth sensors are still imperfect, which might be in low-resolution, noisy, and contaminated with large holes due to reflective surfaces and distant objects in the scene.

\emph{Depth~super-resolution} and~\emph{depth~completion} techniques are designed to overcome these limitations by mainly leveraging the information from high-resolution and sharp color images~\cite{ferstl2013image, yang2014color} to improve the quality of the depth map. Nevertheless, in real world settings, the quality of color images could be significantly variable due to camera vibration and relative motion of the dynamic objects in the scene.

Existing works use multiple-view blurry images to estimate the depth and deblur the image~\cite{arun2015multi, hu2014joint, seok2013dense}. Even though they demonstrate that the depth estimation would also benefit image deblurring, their frameworks cannot be directly adopted to solve our problem, since they make strong assumptions that the scene is static, and the blur is only due to camera shake. In outdoors scenarios, the blur is also generated by the motion of dynamic objects and limited field of depth of the camera. Ismael \etal \cite{al2016real} recently proposed to adopt temporal information to super-resolve depth videos. Their method is constrained to particular types of 3D motion between neighboring frames such as the motion then can be decoupled into a lateral term and radial displacements. Although they attempted depth improvement, their method does not enhance the color image quality.

\begin{figure}
\begin{center}
\begin{tabular}{cc}
\hspace{0.0cm}
\includegraphics[width=0.18\textwidth]{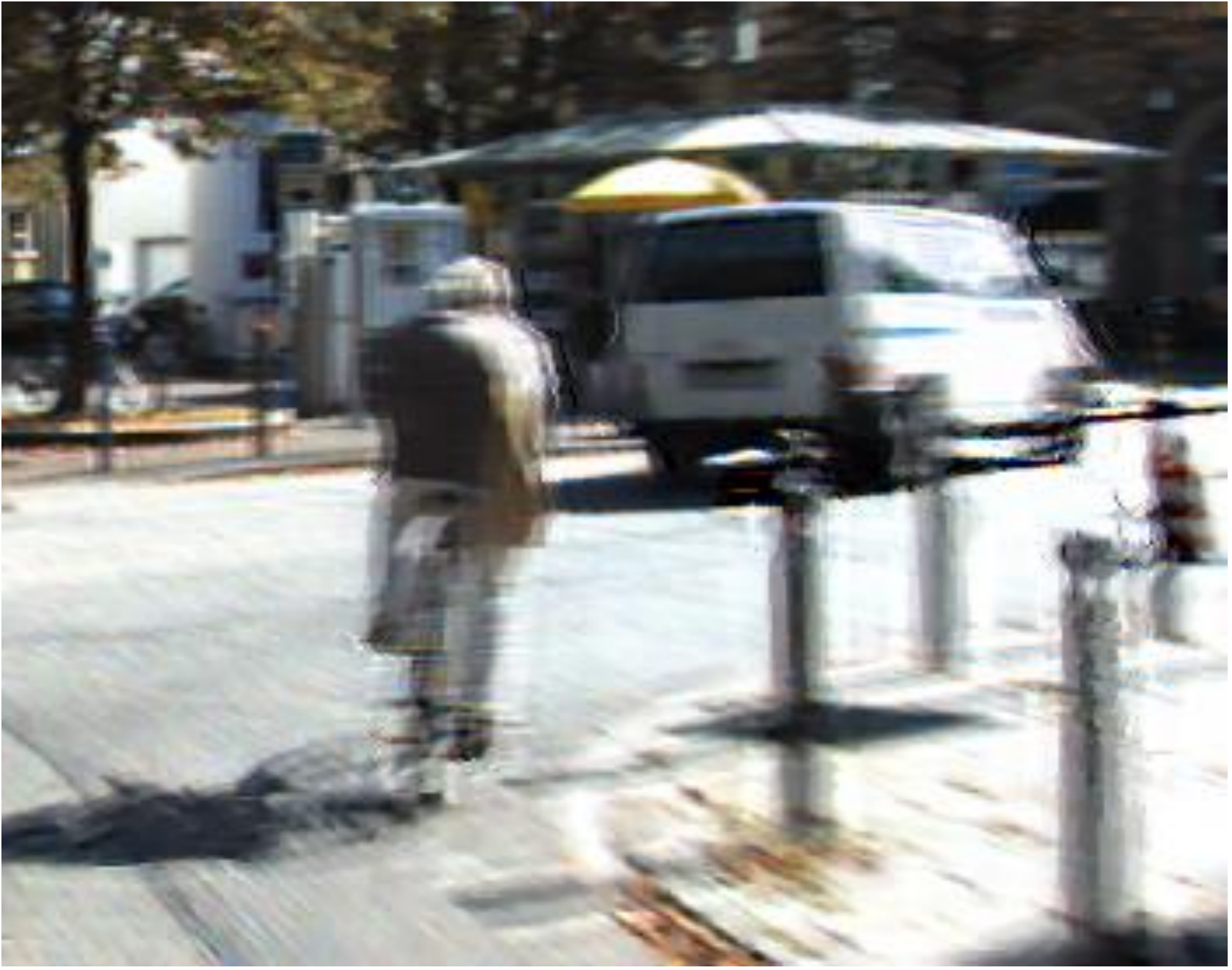}
&\hspace{0.0cm}
\includegraphics[width=0.18\textwidth]{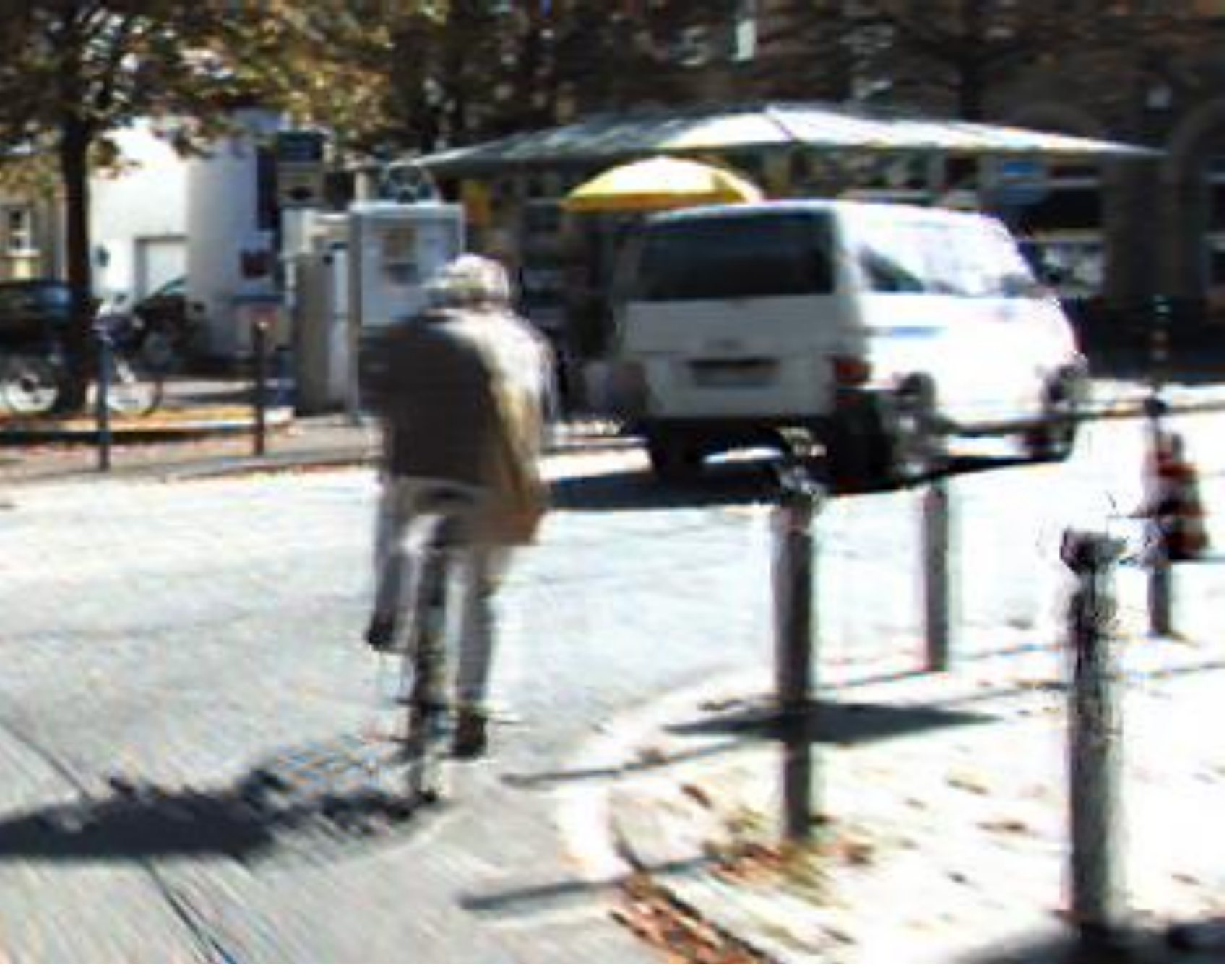}\\
(a) $r = 16$ & (b) $r = 8$ \\
\hspace{0.0cm}
\includegraphics[width=0.18\textwidth]{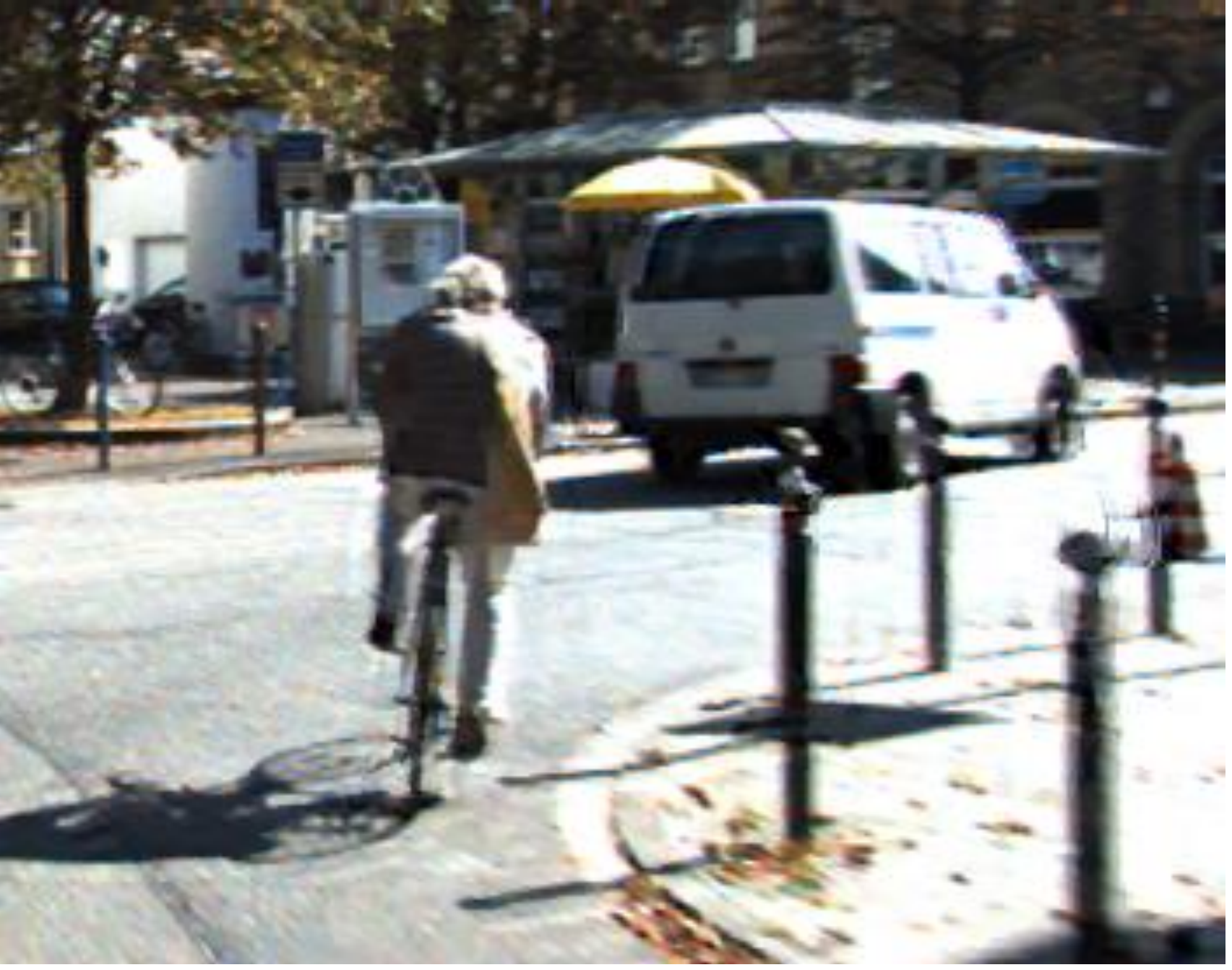}
&\hspace{0.0cm}
\includegraphics[width=0.18\textwidth]{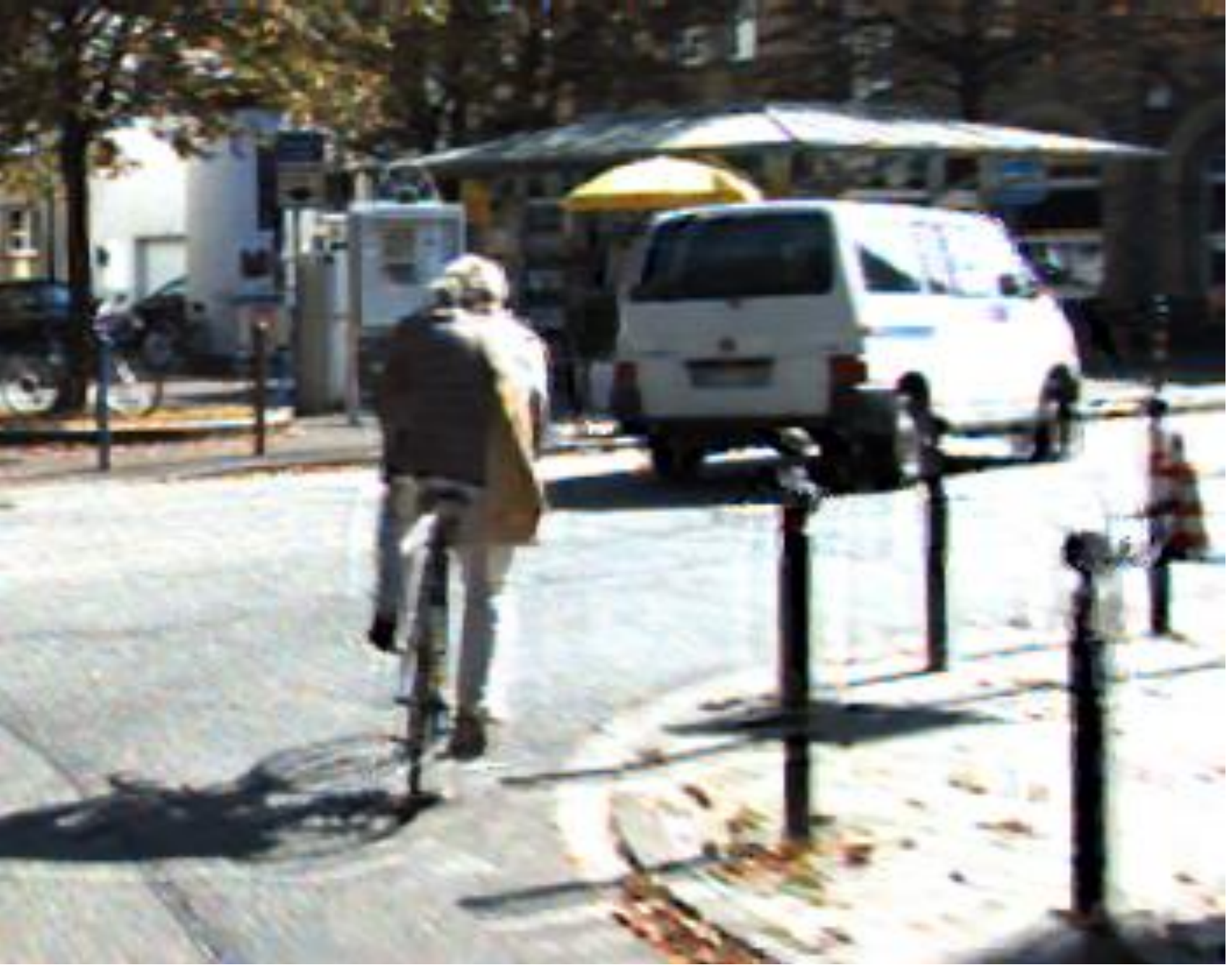}\\
(c) $r = 4$   & (d) $r = 0$ \\
\end{tabular}
\end{center}
\caption{Performance of the image deblurring part of our method. Depth maps at different resolutions where $r$ is the downsampling factor are shown (Best viewed on screen).}
\label{fig:depthcompare}
\end{figure}

In Fig.~\ref{fig:first}(a), we provide a sample outdoor traffic scene image depicting camera and object motions. The undesired blur in the image causes the loss of details, which further hinders depth completion results. As indicated in Table~\ref{blurvsclean}, the performance of the state-of-the-art depth completion methods quickly deteriorate in the presence of blur in color images. On the other hand, the quality of the depth map has significant influence in deblurring color images. In Fig.~\ref{fig:depthcompare}, we compare the deblurring results with different resolution depth maps using our method. We observe that the deblurring performance improves with the increase of depth map quality. Therefore, we conclude that depth map completion and image deblurring are interweaved and strongly co-dependent where the solution of one benefits the other.

In this paper, we focus on handling realistic scenarios and tackling the problem of joint depth map completion and image deblurring by exploiting the spatio-temporal constraints in color images and depth map sequences. Our work is motivated by the recent progress in image deblurring and depth completion. It has been demonstrated~\cite{sellent2016stereo} that scene flow estimation from stereo pairs can significantly improve the deblurring performance. This indicates that depth information can lead to a better deblurring in varying conditions compared to solely image-based methods. Likewise, deblurred images can support depth completion to estimate high-quality depth maps~\cite{al2016real, zhang2009consistent}.

To this end, we introduce a new framework for joint restoration of scene depth map and the latent clean image from given sparse depth maps and their corresponding blur color image sequences. Specifically, we use the piecewise planar assumption of the scene and represent the entire scene as a collection of 3D local planes, which significantly regularizes the problem. In this way, the joint restoration of scene depth map and latent clean image have been transformed to the estimation of the 3D geometry for each local plane, the rigid motion for each plane and the solution for the latent clean image. Our main contributions can be summarized as a comprehensive and efficient energy minimization formulation and the state-of-the-art depth completion performance using multiple images.

\section{Related Work}
\begin{table}\footnotesize
\centering
\caption{Comparisons with clean/blur images on KITTI dataset.
}
\label{blurvsclean}
\begin{tabular}{c|c|c|c|c}
\hline
\multirow{2}{*}{KITTI}                               & \multicolumn{2}{c|}{Flow Error(\%)} & \multicolumn{2}{c}{Depth Error(\%)}  \\ \cline{2-5}
                                               & Clean       & Blur        & Clean        & Blur        \\ \hline
Vogel \etal \cite{vogel20153d}                 & 2.83        & 13.62       & 4.27         & 8.20        \\ \hline
Menze \etal \cite{menze2015object}             & 3.28        & 14.77       & 4.70         & 6.72        \\ \hline
Yang  \etal \cite{yang2014color}               & /           & /           & 3.43         & 4.67     \\ \hline
D Ferstl \etal \cite{ferstl2013image}          & /           & /           & 4.08         & 5.14        \\ \hline
J Park \etal \cite{park2014high}               & /           & /           & 9.76         & 12.61       \\ \hline
\end{tabular}
\end{table}
\vspace{-0.2cm}

Depth map completion has been widely studied in computer vision and image processing, and the research topic can be roughly divided into two categories, namely, depth map only and color image guided.

\noindent\textbf{Depth map only:}
A common way to improve the resolution and quality of depth map is to fuse multiple depth maps into one depth map. KinectFusion \cite{izadi2011kinectfusion} uses depth maps from neighboring frames to fill in the missing information during real time 3D rigid reconstruction. Newcombe \etal \cite{newcombe2011kinectfusion} took live depth data from a moving Kinect camera and created a high-quality 3D model for a static scene. Ismaeil \etal \cite{al2016real} proposed to complete low-resolution dynamic depth videos containing non-rigidly moving objects with a dynamic multi-frame super-resolution approach. This is obtained by accounting for nonrigid displacements in 3D, in addition to 2D optical flow, and simultaneously correcting the depth measurement by Kalman filtering. However, the real challenge that the research community has been facing is extending the multi-frame depth completion concept to dynamic scenes with moving objects.

\noindent\textbf{Color image guided depth completion:}
This category of methods use additional intensity image as guidance for depth completion~\cite{garcia2011new, garcia2015unified, or2015rgbd, zhu2011reliability}. Yang \etal \cite{yang2007spatial} used bilateral filtering of a depth cost volume and a RGB image in an iterative refinement process. A more complex approach was proposed by Park \etal \cite{park2011high, park2014high} to use a combination of different weighting terms of a least squares optimization including segmentation, image gradients, edge saliency and non-local means for depth upsampling. Ferstl \etal\cite{ferstl2013image} modeled the smooth term as a second order total generalized variation (TGV) regularization, and guided the depth upsampling with an anisotropic diffusion tensor calculated from a high-resolution intensity image. However, this method suffers from the blurring problems, especially areas around depth edges. Yang \etal \cite{yang2014color} developed an adaptive color-guided auto-regression model for depth recovery. Aodha \etal \cite{mac2012patch} focused on single image upsampling as MRF labeling problem.

\noindent\textbf{Deep CNN based depth completion:}
Recently, CNN has shown its ability in image recognition and classification task \cite{simonyan2014very,yang2017learning} and has been extended to low-level vision tasks such as depth map super-resolution or depth completion. Riegler \etal~\cite{Riegler:ATGV-Net:ECCV2016} proposed a unified framework to effectively combine DCNN with total variations to generate HR depth maps. Riegler \etal~\cite{riegler:A-deep-primal-dual-netwok-for-guided-depth-super-resolution:BMVC2016} proposed to incorporate non-local variation into DCNN based framework, where the corresponding color images were also utilized. Additionally, Hui \etal~\cite{hui:Depth-map-super-resolution-by-deep-multi-scale-guidance:ECCV2016} proposed a multi-scale guided convolutional network (MSG-Net). All these deep CNN based methods depend on the consistency between the training data and the testing data.

\noindent\textbf{Deblurring with depth :}
Blur removal is an ill-posed problem, thus certain assumptions or additional constraints are required to regularize the solution space. As depth can significantly simplify the deblurring problem, depth-aware methods have been proposed to leverage the depth information. Xu \etal~\cite{xu2012depth} inferred depth from two blurry images captured by a stereo camera and proposed a hierarchical estimation framework to remove motion blur caused by in-plane translation. Hu \etal \cite{hu2014joint} solved it as a segment-wise depth estimation problem by assuming a discrete-layered scene where each segment corresponds to one layer. Arun \etal~\cite{arun2015multi} proposed a geometric algorithm to estimate the camera motion from the blurry images themselves. However, they all assume that the scene to be static and the camera motion is the only source of motion blur. Recently, Sellent \etal~\cite{sellent2016stereo} proposed a stereo deblurring approach, where 3D scene flow is estimated from the blurry images using a piecewise rigid 3D scene flow \cite{vogel20153d} representation. Very recently Pan \etal ~\cite{Pan_2017_CVPR} proposed a single framework to jointly estimate the scene flow and deblur the images, where the motion cues from scene flow estimation and blur information could reinforce each other. Inspired by this stereo deblurring work, we aim to use a single view image sequence and its sparse and noisy depth map to complete the depth map and estimate the latent clean images.

\section{Problem Formulation}\label{sec:model}


Our goal is to complete the given incomplete and noisy depth maps $\tilde{\mathbf{D}}$ with the help of blurry color images $\mathbf{B}$ by exploiting the spatial-temporal constraints. Blur is caused by the motion of camera, objects, and limited depth-of-field of the camera (for large depth variations in the scene).

Towards this goal, we formulate our problem as a joint depth map completion and color image deblurring under dynamic scene settings. Since there are more variables (latent clean color images and target completed depth maps) to infer than the available measurements (blurry color images and incomplete and noisy depth measurements), we regularize this under-determined problem with the assumption that the scene can be well approximated by a collection of 3D planes~\cite{yamaguchi2013robust} belonging to a finite number of objects performing rigid motions~\cite{menze2015object}, \ie a piecewise planar rigid motion representation. In this way, the original problem is transformed to the estimation of the geometric parameters $\mathbf{n}_i$, the local rigid motion $(\mathbf{R}_i,\mathbf{t}_i)$ for each 3D planar and the latent clean images $\mathbf{I}$.

In the following sections, we describe how to combine the geometric parameters, the motion parameters and the latent clean images together in the same objective where the solution of one variable will benefit the other variables. Our model relates to \cite{Pan_2017_CVPR} and \cite{menze2015object} in estimating the scene flow. However, our problem setting is very different where we have to exploit the sparse depth measurement constraint and blurry image constraint in a joint framework. We have introduced two new depth constraints to evaluate the consistency and the discrepancy between the sparse depth measurements and the completed depth maps.

\subsection{Blur Image Formation}
\label{sec:3.1}
For complex dynamic settings such as outdoor traffic scenes, the blurry image is generated by spatially-variant per-pixel motion (optical flow). The blurry images are formed by the integration of light intensity emitted from the dynamic scene over the aperture time interval of the camera,
\begin{equation} \label{eq:convBlurKernel}
\mathbf{B}_m({\bf x}) =\frac{1}{2N+1} \sum^{N}_{n=-N} \mathbf{I}_n({\bf x}+{\bf u}_n) = {\bf A}_m^x{\bf I}_m(x),
\end{equation}
where ${\bf B}$ is the blurry frame, $\mathbf{x}$ denotes pixel location on image domain,  $\mathbf{I}_n$ is the successive latent neighboring frames as frame $m$, $\mathbf{u}_n$ is the optical flow to frame $n$, $\mathbf{A}_m^x$ is the blur kernel vector for the image at location $\mathbf{x}$. We obtain the blur kernel matrix $\mathbf{A}$ by stacking $\mathbf{A}^x$. This leads to the blur model for the image as $\mathbf{B}_m = \mathbf{A}_m{\bf I}_m$. Please refer to \cite{hyun2015generalized} and \cite{sellent2016stereo} for more details.
\subsection{Formation Statement}
In our setup, the incomplete and noisy depth measurements provide the depth information for each frame. Based on our piece-wise rigid planar assumption of the scene, optical flows for pixels lying on the same plane are constrained by the same homography. In particular, we represent the scene in terms of superpixels and finite number of objects with rigid motions. We denote $\mathcal{S}$ and $\mathcal{O}$ as the set of superpixels and moving objects, respectively. Each superpixel $i\in S$ is associated with a region $\mathcal{R}_i$ in the image and a plane variable $\mathbf{n}_{i,k} \in \mathbb{R}^3$ in 3D ($\mathbf{n}_{i,k}^T\mathbf{X}=1$ for $\mathbf{X} \in \mathbb{R}^3$), where $k \in \left\{1,\cdots,|\mathcal{O}|\right\}$ denotes superpixel $i$ is associated with the $k$-th rigid motion $\mathbf{o}_k=(\mathbf{R}_k, \mathbf{t}_k) \in \mathbb{SE}(3)$, where $\mathbf{R}_k \in \mathbb{R}^{3 \times 3}$ is the rotation matrix and $\mathbf{t}_k \in \mathbb{R}^{3}$ is the translation vector. Given the motion parameters ${\mathbf o}_k$ and geometric parameters $\mathbf{n}_{i,k}$, we can obtain the homography defined for superpixel $i$ as
\begin{equation}
\mathbf{H}(\mathbf{n}_{i},\mathbf{o}_{k}) = {\mathbf K}(\mathbf{R}_k -\mathbf{t}_k \mathbf{n}^{T}_{i,k}){\mathbf K}^{-1},
\end{equation}
where ${\mathbf K} \in \mathbb{R}^{3 \times 3}$ is the intrinsic calibration matrix. ${\mathbf H}$ for each superpixel can be obtained when relates correspondences across frames $\mathbf{n},\mathbf{o}$ are confirmed.
This shows that the optical flows for pixels in a same superpixel are constrained by the corresponding homography, thus the optical flows are structured as opposed to \cite{hyun2015generalized}.

We aim at completing the incomplete and noisy depth maps by exploiting both the spatial-temporal information in constraining the motion and the availability of corresponding blurry color images. To this end, we formulate the problem in a single framework as a discrete-continuous optimization problem to jointly complete the depth maps and deblur the color images. We explain all the constraints in the following sections.

\subsection{Depth Constraint}
\label{sec:depthTerm}
\subsubsection{Depth Consistency}
The first depth term is to encourage the consistency between the sparse depth measurements and the completed depth estimation based on the piecewise planar models, which are evaluated across multiple frames. For the reference frame, the depth consistency is defined as:
\begin{equation}
\psi_i^1(\mathbf{n}_{i,k}) = w_{1} \sum_{\mathbf{x} \in \Omega} | \tilde{\mathbf D}(\mathbf x)-{\mathbf D}(\mathbf{n}_{i,k},\mathbf x)|_1,
\end{equation}
where $\tilde{\mathbf D}(\mathbf x)$ denotes the sparse and noisy depth measurements from sensors such as Kinect and LiDAR, $\Omega$ denotes the image pixels with depth measurements available and ${\mathbf D}(\mathbf{n}_{i,k},\mathbf x)$ represents the depth estimation under the piecewise planar model.

For other frames besides the reference frame, the second depth consistency is evaluated as the discrepancy between the measured depth and the corresponding depth generated with the piecewise rigid planar motion,
\begin{equation}
\psi_i^2(\mathbf{n}_{i,k},\mathbf{o}_{k}) = w_{2}\sum_{\mathbf{x} \in \Omega}|\mathbf D(\mathbf x,\mathbf{n}_{i,k},\mathbf{o}_{k}) - \tilde{\mathbf D} (\mathbf{H}^{*}\mathbf x)|_1,
\end{equation}
where the superscript $*$ denotes the warping direction to other color frames and the subscript of $\mathbf{H}$ is to index the corresponding homography for position $\mathbf{x}$.
\subsubsection{Motion Sensitive Depth Discontinuity}
Our model exploits a smoothness potential that enforcing the depth maps to be smooth and continuous, which involves both discrete and continuous variables. It is similar to the ones used in~\cite{menze2015object}. The third depth term for depth map is to enforce the motion boundaries to be co-aligned with the depth discontinuities, which is expressed as
{\small
\begin{eqnarray}
\hspace{-0.8cm}&&\psi_{i,j}^3(\mathbf{n}_{i,k},\mathbf{n}_{j,k'}) \nonumber\\
\hspace{-0.8cm}&&=w_{3}\left\{\begin{tabular}{cc}
$\hspace{-0.1cm}\exp {\Big(}-\frac{\lambda}{|\mathcal{B}_{i,j}|}\sum\limits_{\mathbf{x} \in \mathcal{B}_{i,j}} \omega_{i,j}(\mathbf{n}_i,\mathbf{n}_j,\mathbf{x})^2 \frac{|\mathbf{n}_i^T \mathbf{n}_j|}{\left\| \mathbf{n}_i \right\| \left\| \mathbf{n}_j \right\|}{\Big)}$& if  $k \neq k'$,\nonumber\\
0 & else.
\end{tabular}
\right.
\end{eqnarray}
}
${\mathbf x} \in \mathcal{B}_{i,j}$ evaluated with $i$-th superpixel parameter and $j$-th superpixel parameter, where $|\mathcal{B}_{i,j}|$ denotes the number of pixels belongs to the boundary between superpixels $i$ and $j$.
\subsubsection{Geometry Sensitive Depth Smoothness}
The fourth depth term is to enforce the compatibility of two superpixels that share a common boundary by respecting the depth discontinuities. We define our potential function for continuous boundary as
{\small
\begin{equation}
\begin{aligned}
\psi_{i,j}^4(\mathbf{n}_{i,k},\mathbf{n}_{j,k'})  & = \sum_{\mathbf{x}\in \mathcal{B}_{i,j}}\rho_{\alpha_1}(d(\mathbf{n}_{i,k},\mathbf{x})-d(\mathbf{n}_{j,k'},\mathbf{x}))\\
& + \rho_{\alpha_3} \left(1-\frac{|\mathbf{n}_{i,k}^T \mathbf{n}_{j,k'}|}{\left\| \mathbf{n}_{i,k} \right\| \left\| \mathbf{n}_{j,k'} \right\|}\right),
\end{aligned}
\end{equation}
}
where $\rho_{\alpha}(\cdot)=\min(|\cdot|,\alpha)$ denotes the truncated $\ell_1$ penalty function.

\begin{figure*}
\begin{center}
\includegraphics[width=0.80\textwidth]{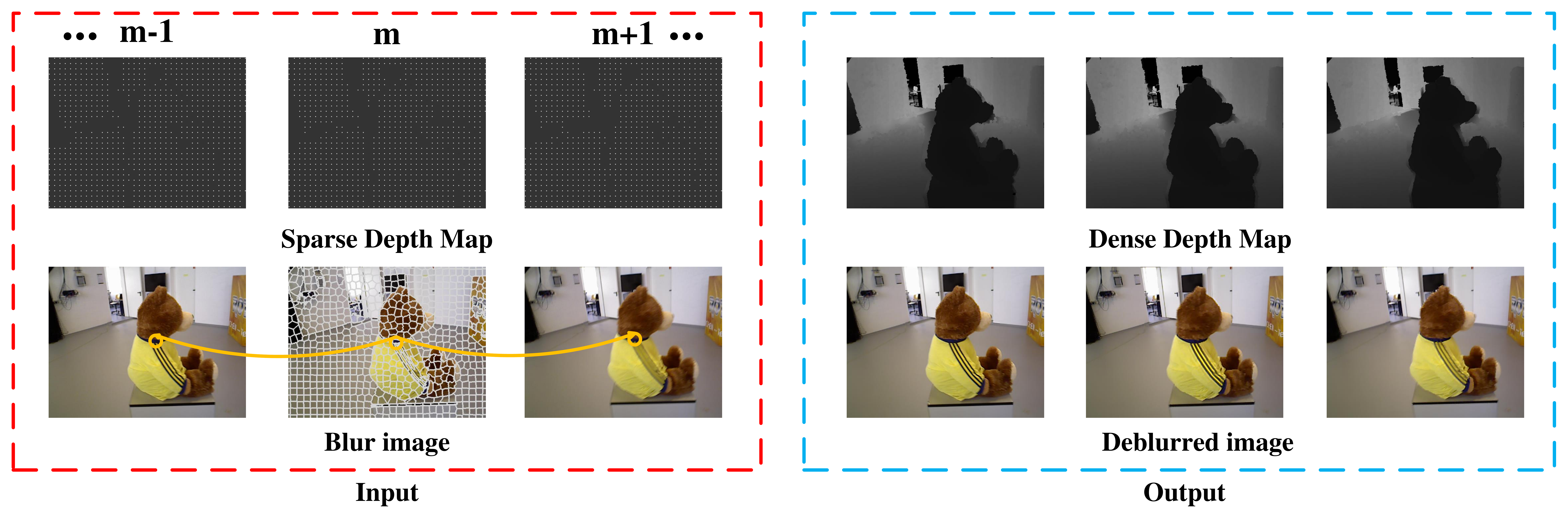}
\end{center}
\caption{Illustration of our method. We simultaneously complete the depth maps and deblur the color images.}
\label{fig:framework}
\end{figure*}

\subsection{Image Constraint}
\label{sec:ImageTerm}
\subsubsection{Brightness Consistency}
Our image term involves mixed discrete and continuous variables, and are of three different kinds. The first image term encourages the corresponding pixels across the latent clean images should own similar appearance,
\begin{equation}
\theta_i^1(\mathbf{n}_{i},\mathbf{o},\mathbf{I}) = c_{1}|\mathbf I(\mathbf x)-\mathbf I^{*}(\mathbf H^{*}\mathbf x)|_1,
\end{equation}
where $\mathbf I^{*} \in \mathbf{I}_n$ denotes the frame which $\mathbf{H}^*$ warping to. We use the $\ell_1$ norm due to its robustness against noise and occlusions.
\subsubsection{Anchor Point Constraint}
While the above brightness consistency term provides dense constraint across image frames, we could also exploit the sparse and reliable feature correspondences (such as SIFT) to constrain the correspondences, which work as anchor points. Therefore our second image term is defined as
{\small
\begin{equation}
\theta_i^2(\mathbf{n}_{i},\mathbf{o})=\left \{
  \begin{tabular}{cc}
  $c_2$$\rho_{\alpha_2}(||{\mathbf H}^{*}{\mathbf x}-\mathbf{x}^{'}||_2)$& if $\mathbf{x}\in \Pi$ \\
  0 & otherwise. \nonumber \\
  \end{tabular}
\right.
\end{equation}
}
More specifically, it encodes the information that the warping of feature points $\mathbf{x}\in \Pi$ based on ${\mathbf H}^{*}$ should match its extracted correspondences $\mathbf{x}^{'}$ in the target view.
\vspace{-0.2cm}
\subsubsection{Deblurring Constraint}
The third image term relates the observed blurry images with the latent clean images with the spatial-variant blur kernels,
{\small
\begin{equation}
\mathbf{\theta}_i^3(\mathbf{n}_{i},\mathbf{o},\mathbf{I})= c_{3}\sum_m\sum_{\partial}\left\|{\partial}\mathbf{A}_{m}(\mathbf{n}_{i},\mathbf{o})\mathbf{I}_m - {\partial}\mathbf{B}_m\right\|_2^2 \nonumber
\end{equation}
}
where $\partial(\cdot)$ denotes the Toeplitz matrices corresponding to the horizontal, vertical derivative filters and the identity matrix. This term encourages the intensity changes and the intensity in the estimated blurry images to be close to that of the observed blurry images.

\subsection{Regularization Term for Latent Clean Images}
\label{sec:regularization}
Natural images of typical real-world scenes generally obey sparse spatial gradient distributions \cite{krishnan2009fast, krishnan2011blind}. The distribution of a latent clean image can often be modeled as a generalized Laplace distribution \cite{Jiaolong:CVPR-2016}, \ie
$P(\mathbf{I})= \prod_{\mathbf{x}\in\mathbf{X}}\exp(-|\nabla_{\mathbf{x}}\mathbf{I} (\mathbf{x})|^p),
$
where the power of $p$ is a parameter usually within $[0.0,1.0]$. This prior can be equivalently represented in energy minimization form, \ie $\|\nabla_{\mathbf{x}}\mathbf{I}(\mathbf{X})\|^p\rightarrow \min.$
We let $p=1$ in the paper. In our model, this corresponds to a total variation term to suppress the noise in the latent image while preserving edges, and penalize spatial fluctuations.
\begin{equation} \label{Eregularization1}
\psi_m = |\nabla \mathbf{I}_m| = \|\mathbf{I}_m\|_{\mathrm{TV}}.
\end{equation}

\vspace{-0.2cm}
\subsection{Energy Minimization}
Our energy minimization is defined as
{\small
\begin{eqnarray}\label{eq:energy}
\begin{aligned}
E& = \underbrace{\sum_{i\in \mathcal{S}}\mathbf{\psi}_{i}^{1,2}(\mathbf{n}_{i},\mathbf{o}) + \sum_{i,j\in \mathcal{S}}\mathbf{\psi}_{i,j}^{3,4}(\mathbf{n}_{i},\mathbf{n}_{j},\mathbf{o}) }_{\text{depth map}} \\
& + \underbrace{\sum_{i\in \mathcal{S}} \mathbf{\theta}_{i}^{1,2,3}(\mathbf{n}_{i},\mathbf{o},\mathbf{I})}_{\text{image~term}} + \underbrace{\sum_m\mathbf{\psi}_{m}(\mathbf{I}_m)}_{\substack{\text{clean image}\\\text{regularisation}}},
\end{aligned}
\end{eqnarray}
}
which consists of data terms evaluated on the color images and depth maps respectively, a smoothness term for the desired completed depth map, and a spatial regularization term for the latent clean images. Our model has been defined on three consecutive frames of RGB-D sequences. It can also allow the input with two pairs of RGB-D frames. Details are provided in section~\ref{sec:experiments}.  In Section~\ref{sec:optimization}, we perform the optimization in an alternating manner to handle mixed discrete and continuous variables, thus allowing us to jointly complete the depth maps, and deblur the color images.


\section{Solution of Energy Function}\label{sec:optimization}
The optimization of our energy function defined in Eq.-(\ref{eq:energy}) involves both discrete and continuous variables, which is challenging to solve. Therefore we resort to the alternative optimization manner, \ie, optimizing one variable while fixing all the remaining ones.  Note that our energy minimization formulation involves three set of variables, namely completed depth map $\mathbf{D}$ as indexed by $\mathbf{n}$, piecewise planar rigid motion $\mathbf{o}$ and latent clean image $\mathbf{I}$. We propose to handle the energy minimization by alternating between the following two steps,
\begin{itemize}
\item Fix the latent clean image ${\mathbf I}$, solve for scene geometry $\mathbf{n}$ and motion $\mathbf{o}$ (completed depth map and motion) by optimizing Eq.(\ref{eq:sceneFlowEnergy}) (See Section~\ref{sec:sceneflow}).
\item Fix the scene geometry and motion ${\mathbf n}$ and ${\mathbf o}$, solve for the latent clean image $\mathbf{I}$ by Eq.(\ref{eq:latentImageEnergy}) (See Section~\ref{sec:deblurring}).
\end{itemize}

\subsection{Depth Completion and Motion Estimation}\label{sec:sceneflow}
When the latent clean images are fixed as $\mathbf{I} = \tilde{\mathbf{I}}$, the joint optimization in  Eq.(\ref{eq:energy}) reduces to
{\small
\begin{equation}\label{eq:sceneFlowEnergy}
\min_{{\mathbf n},{\mathbf o}}\sum_{i\in \mathcal{S}}\mathbf{\psi}_{i}^{1,2}(\mathbf{n}_{i},\mathbf{o})+ \sum_{i,j\in \mathcal{S}}\mathbf{\psi}_{i,j}^{3,4}(\mathbf{n}_{i},\mathbf{n}_{j},\mathbf{o}) + \mathbf{\theta}_{i,j}(\mathbf{n}_{i},\mathbf{n}_{j},\mathbf{o},\tilde{\mathbf{I}}),
\end{equation}
}
which is a discrete-continuous CRF optimization problem. We use the sequential tree-reweighted message passing (TRW-S) method in \cite{menze2015object} to find an approximate solution.

\subsection{Debblurring}\label{sec:deblurring}
Given the scene geometry $\tilde{\mathbf{n}}$ and motion parameters $\tilde{\mathbf{o}}$, the blur kernel matrix ${\mathbf A}_m$ is derived based on Eq.(\ref{eq:convBlurKernel}). The objective function in Eq. (\ref{eq:energy}) becomes convex w.r.t. $\mathbf{I}$
{\small
\begin{equation}\label{eq:latentImageEnergy}
\min_{{\mathbf I}}\sum_{i\in \mathcal{S}} \mathbf{\theta}_{i}^1(\tilde{\mathbf{n}}_{i},\tilde{\mathbf{o}},\mathbf{I}) +\mathbf{\theta}_{i}^3(\tilde{\mathbf{n}}_{i},\tilde{\mathbf{o}},\mathbf{I})+\sum_m\mathbf{\psi}_{m}(\mathbf{I}).
\end{equation}
}
In order to obtain the latent clean image $\mathbf{I}$, we adopt the conventional convex optimization method \cite{chambolle2011first} and derive the primal-dual updating scheme as follows
{\small
\begin{equation}
\left\{
\begin{gathered}
\begin{aligned}
& \mathbf{p}_{r+1}=\frac{\mathbf{p}_{r}+\gamma \nabla \mathbf{I}_r}{\mathbf{max}(1,\mathbf{abs}(\mathbf{p}_{r}+\gamma\nabla \mathbf{I}_r))}\\
& \mathbf{q}_{r+1}=\frac{\mathbf{q}_{r}+\mu (\mathbf{I}_r-\mathbf{I}_{r}^{*})}{\max(1,\mathbf{abs}(\mathbf{q}_{r}+\mu (\mathbf{I}_r-\mathbf{I}_{r}^{*}))}\\
& \mathbf{I}_{r+1} =\arg \min_{\mathbf{I}} \sum_i c_{3}\sum_{\partial }\left\|{\partial}\mathbf{A}\mathbf{I} - {\partial}\mathbf{B}\right\|_2^2 +\\
& \frac{\left\|[\mathbf{I}-\eta ((\nabla {\bf p}_{r+1})^{T}+\eta({\bf q}_{r+1}-{\bf q}_{r+1}^{*})^{T})]-\mathbf{I}_r\right\|^2}{2\eta}
\end{aligned}
\end{gathered}\right.
\end{equation}
}
where $\mathbf{p}$, $\mathbf{q}$ are the dual variables, $\gamma$,$\mu$ and $\eta$ are the step variants which can be modified at each iteration, and $r$ is the iteration number.


\section{Experiments}\label{sec:experiments}
We evaluate the performance of our method on both outdoors settings and indoors environments. For outdoors evaluation, we use the KITTI \cite{geiger2013vision} autonomous driving benchmark dataset that provides monocular color images along with sparse depth maps from the LiDAR for validation. For indoors scenarios, we use the TUM RGB-D dataset~\cite{sturm2012benchmark} captured with a Kinect sensor. We present and discuss our results on both datasets in the following sections.

\subsection{Experimental Setup}
\noindent{\bf Initialization.} Our model in Section~\ref{sec:model} is formulated on three consecutive RGB-D images. In particular, we treat the middle frame as the reference image. We adopt the simple linear iterative clustering (SLIC)~\cite{achanta2012slic} to generate the superpixels, where each superpixel corresponds to a local planar in 3D. We use the penalized least squares method~\cite{garcia2010robust} to fast smooth the given sparse depth map for initialization. The motion hypothesis are then generated using RANSAC algorithm as suggested in~\cite{geiger2011stereoscan}.

\noindent{\bf Evaluations.} Since our method could simultaneously complete the depth map and deblur the given images, we evaluate these two subtasks individually. We evaluate the depth completion results by counting the number of bad pixels having errors more than $3$ pixels and $5\%$ of its ground-truth. We adopt the PSNR to evaluate the deblurring performance. Thus, for each sequence, we report three performance metrics: depth errors (geometry), flow errors (motion), and PSNR (latent images) values for the reference images.

\noindent\textbf{Baselines Methods.}
As for our depth completion results, we compare both passive and active RGB-D methods separately. For multi-view cameras system, we compare with piece-wise rigid scene flow method (PRSF)~\cite{vogel20153d}, which is ranked the firston the KITTI scene flow estimation benchmark and is used as the flow initialization for ~\cite{sellent2016stereo}. For active depth sensors, we compare with TGV~\cite{ferstl2013image}, \cite{park2014high} and \cite{yang2014color} which are also three applicable state-of-the-art methods.
We compare our deblurring results with the state-of-the-art deblurring approach for monocular images~\cite{hyun2015generalized}, and the approach for stereo images~\cite{sellent2016stereo}. The results are shown in Table ~\ref{all_all}.

\begin{table}\footnotesize
\centering
\caption{Quantitative depth completion errors and deblur results. 
}
\label{all_all}
\begin{tabular}{c|c|c|c|c}
\hline
\multirow{2}{*}{}                                    & \multicolumn{2}{c|}{Depth Error(\%)}   & Flow Error(\%)        & PSNR (dB)       \\ \cline{2-5}
                                           & KITTI     & TUM       & KITTI       & KITTI       \\ \hline
Kim and Lee \cite{hyun2015generalized}     & /         & /         & 38.89       & 28.25       \\ \hline
Sellent \etal \cite{sellent2016stereo}     & 8.20      & /         & 13.62~\cite{vogel20153d}  & 27.75       \\ \hline
Yang  \etal \cite{yang2014color}           & 4.67      & 0.43      & /           & /           \\ \hline
D Ferstl \etal \cite{ferstl2013image}      & 5.14      & 0.47      & /           & /           \\ \hline
J Park \etal \cite{park2014high}           & 12.61     & 0.29      & /           & /           \\ \hline
Ours(no depth term)                        &  5.53     & 0.26      &  17.16    & 29.85     \\ \hline
Ours                                       & \bf3.91   & \bf0.22   & \bf13.01    & \bf29.83     \\ \hline
\end{tabular}
\end{table}


\subsection{Experimental Results}

\begin{figure*}[h]
\begin{center}
\begin{tabular}{cccc}
\hspace{0.0cm}
\includegraphics[width=0.243\textwidth]{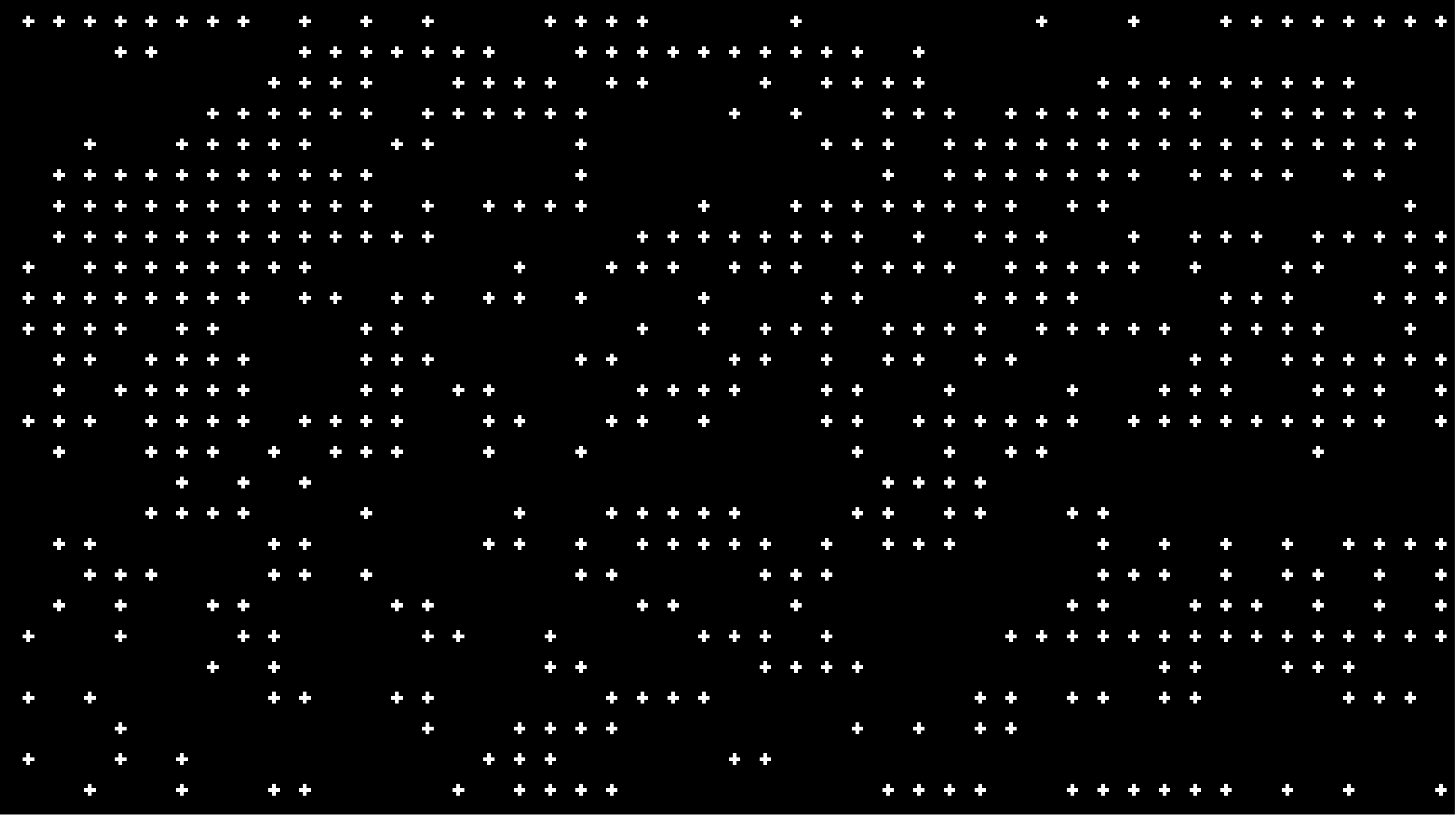}
&\hspace{0.0cm}
\includegraphics[width=0.189\textwidth]{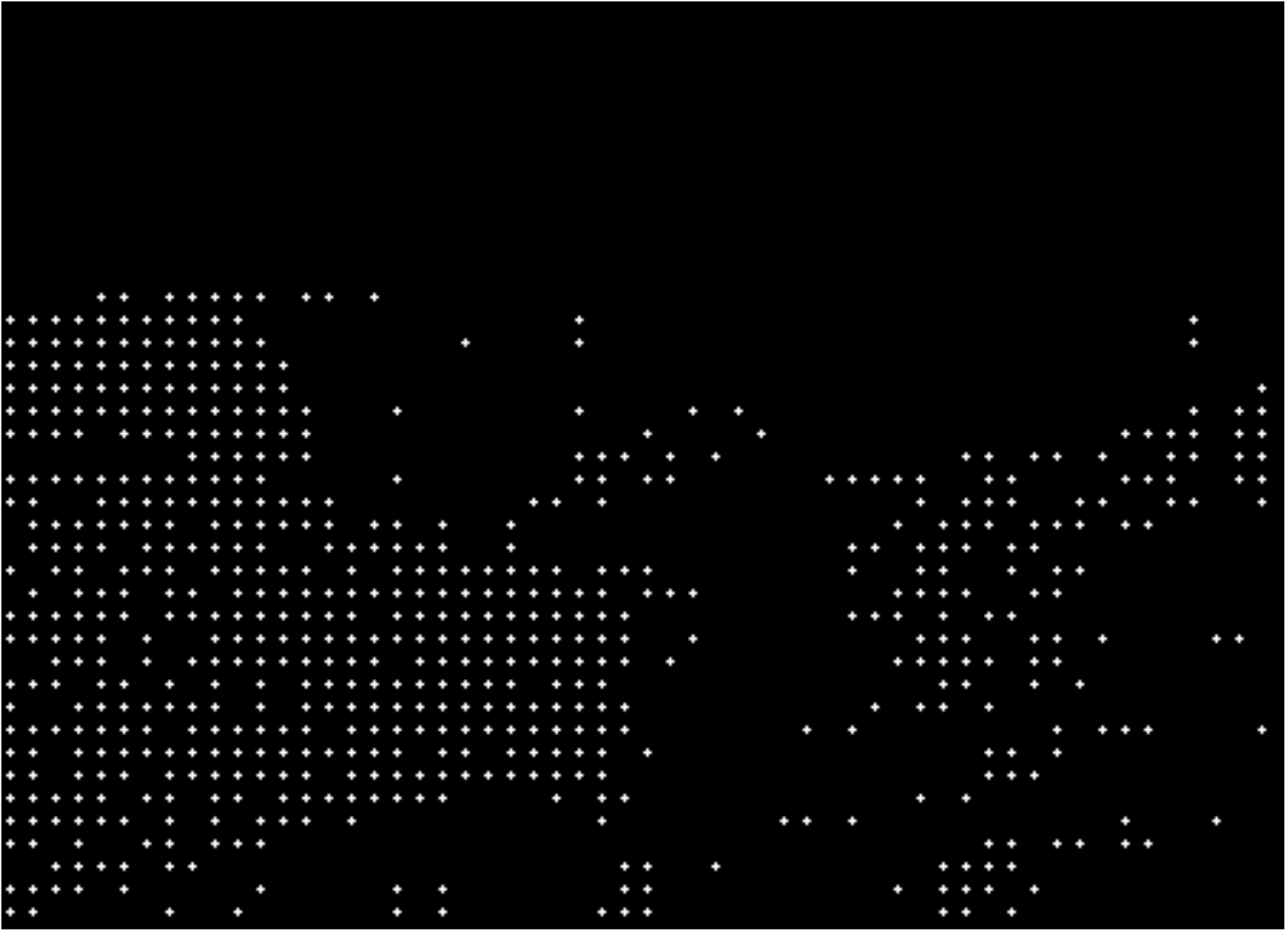}
&\hspace{0.0cm}
\includegraphics[width=0.243\textwidth]{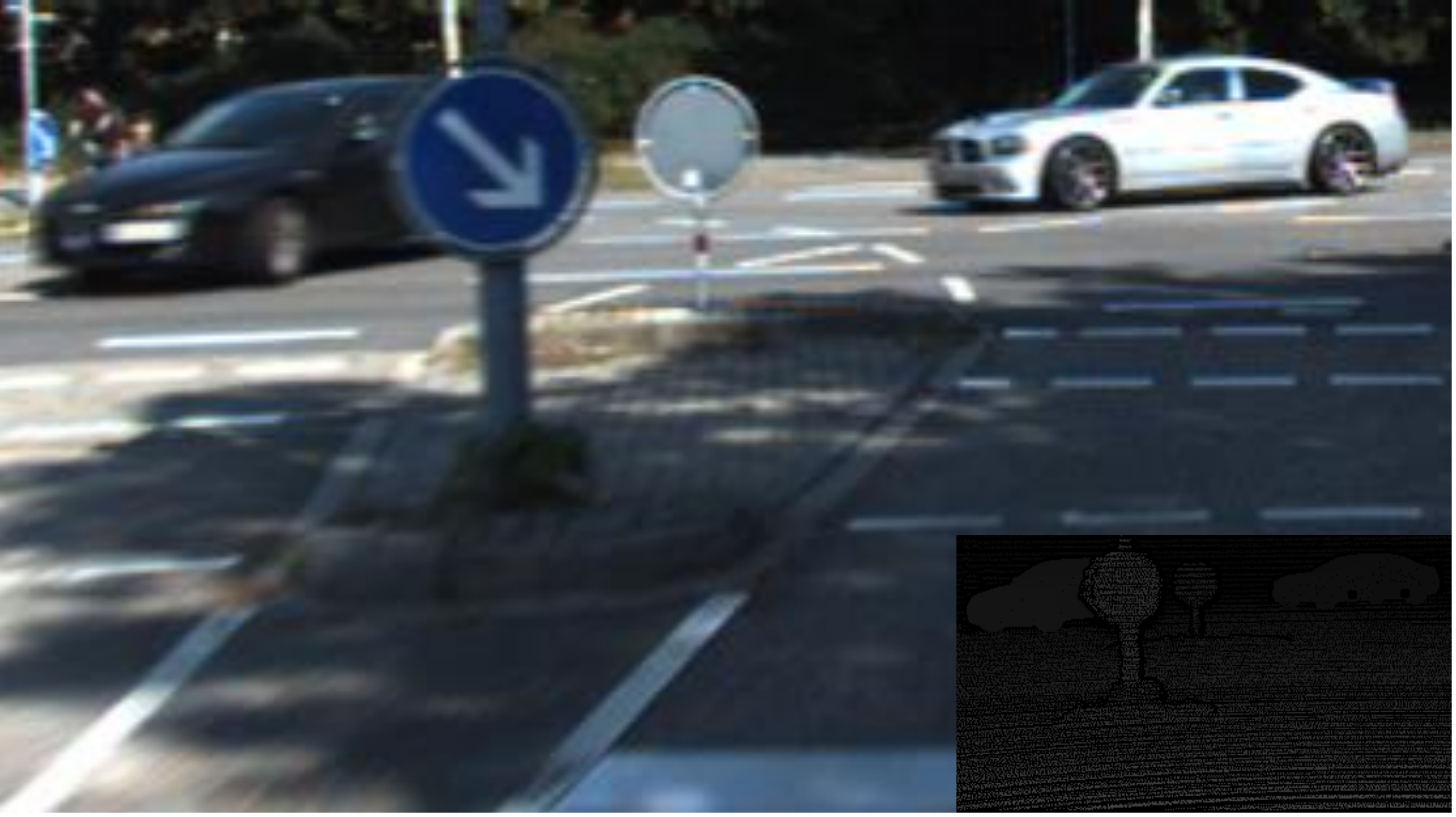}
&\hspace{0.0cm}
\includegraphics[width=0.192\textwidth]{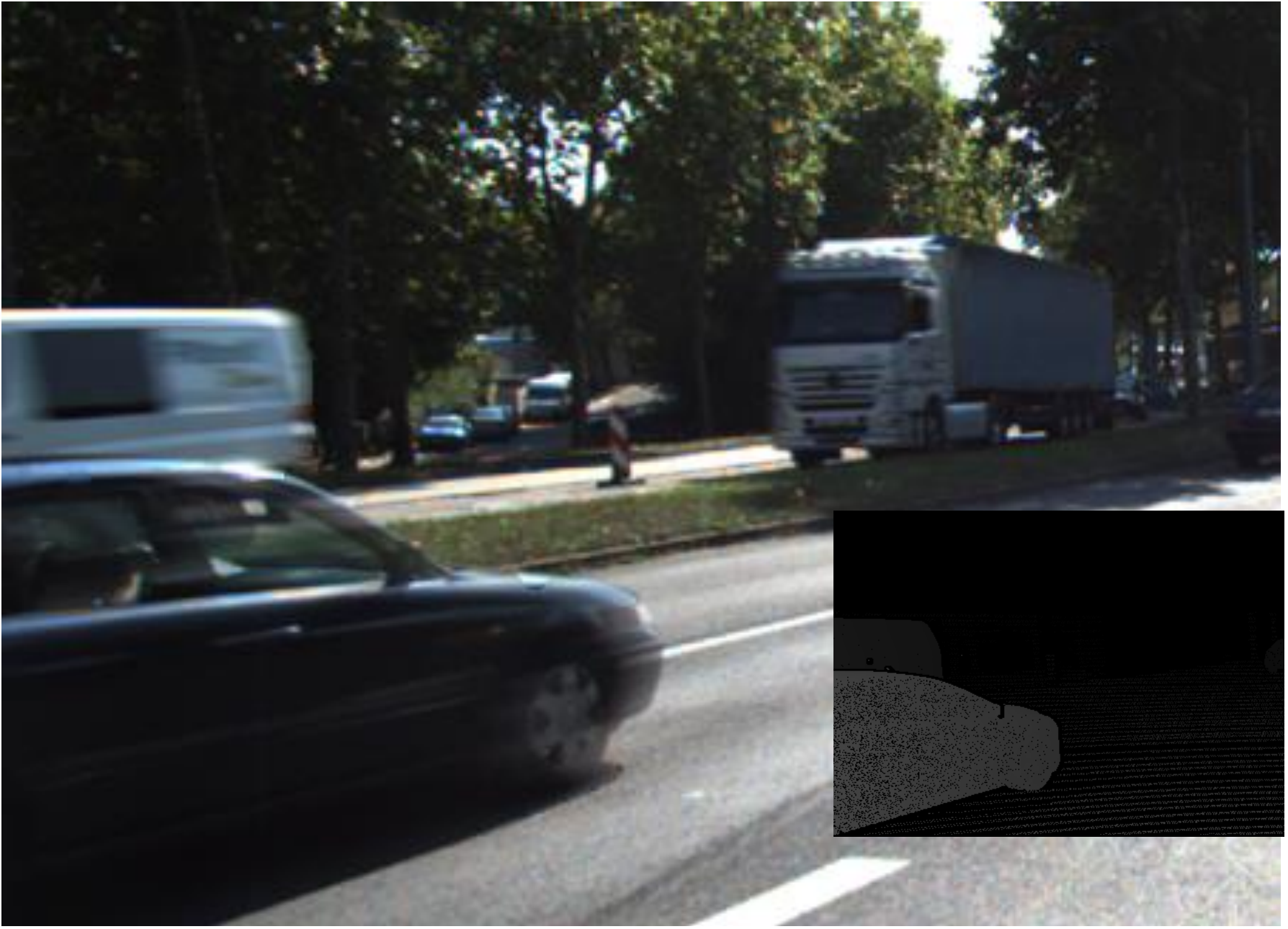}\\
\multicolumn{2}{c}{(a) Input sparse depth map} & \multicolumn{2}{c}{(b) Input blurry image}  \\
\hspace{0.0cm}
\includegraphics[width=0.243\textwidth]{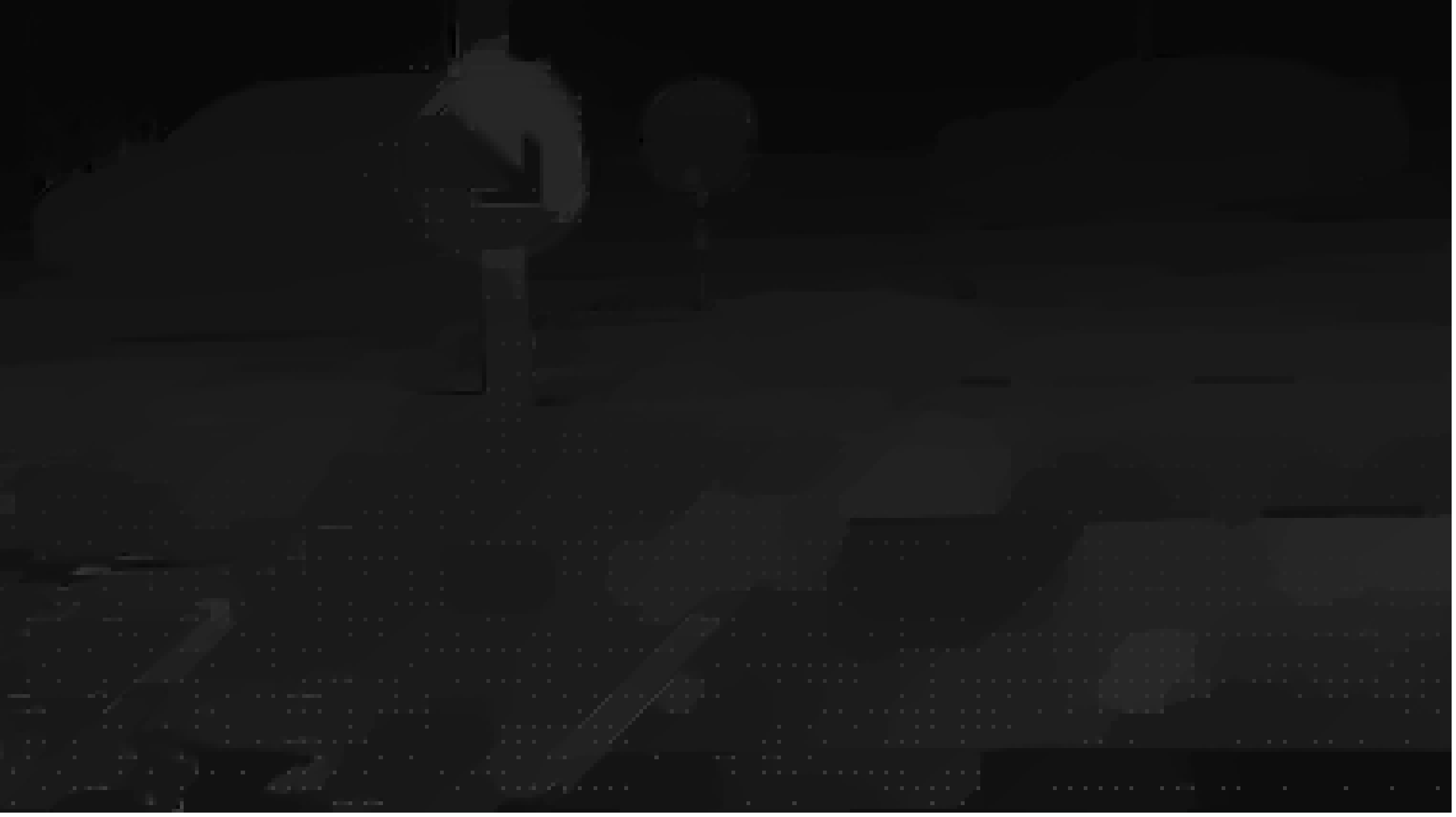}
&\hspace{0.0cm}
\includegraphics[width=0.189\textwidth]{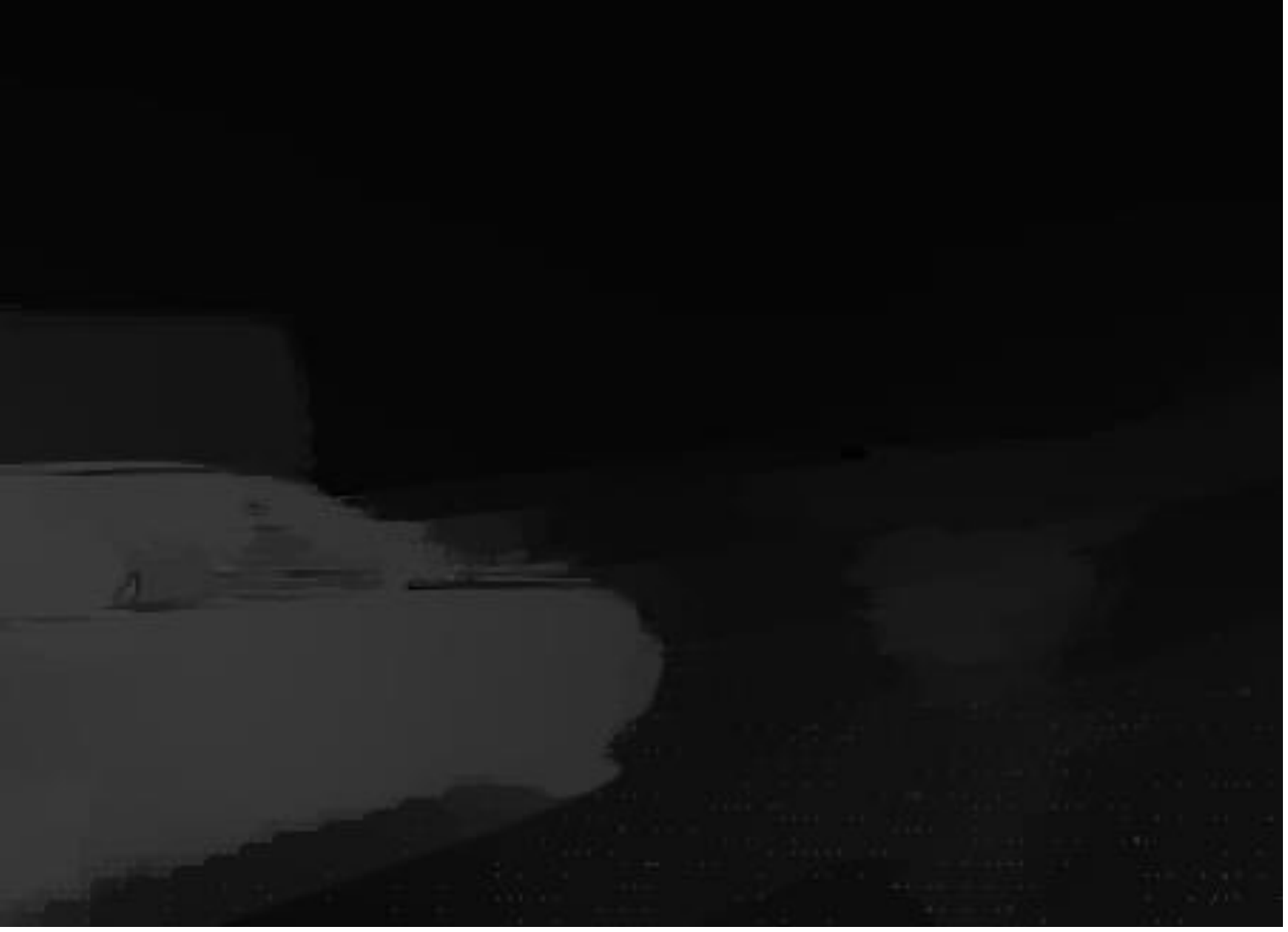}
&\hspace{0.0cm}
\includegraphics[width=0.243\textwidth]{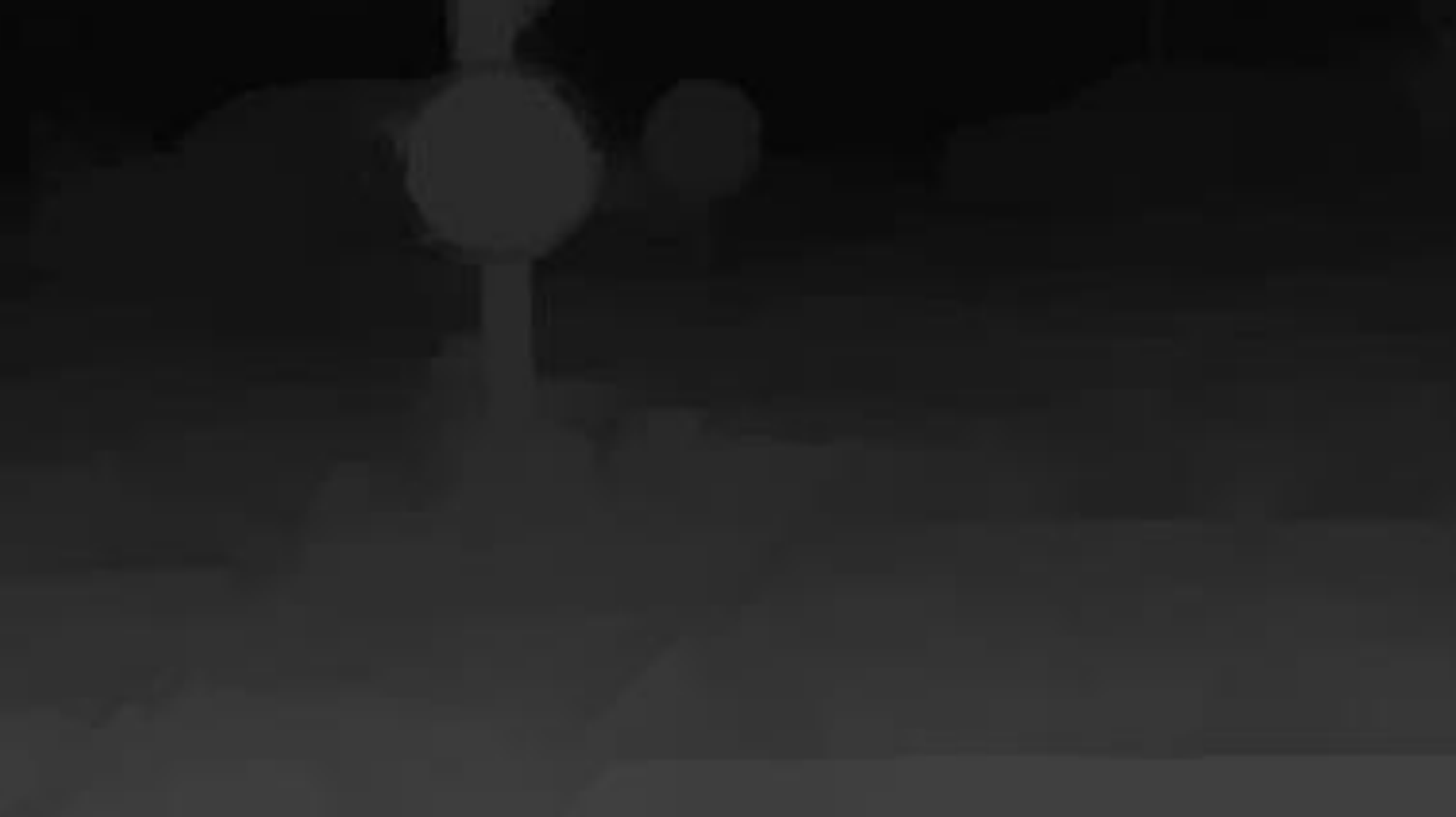}
&\hspace{0.0cm}
\includegraphics[width=0.189\textwidth]{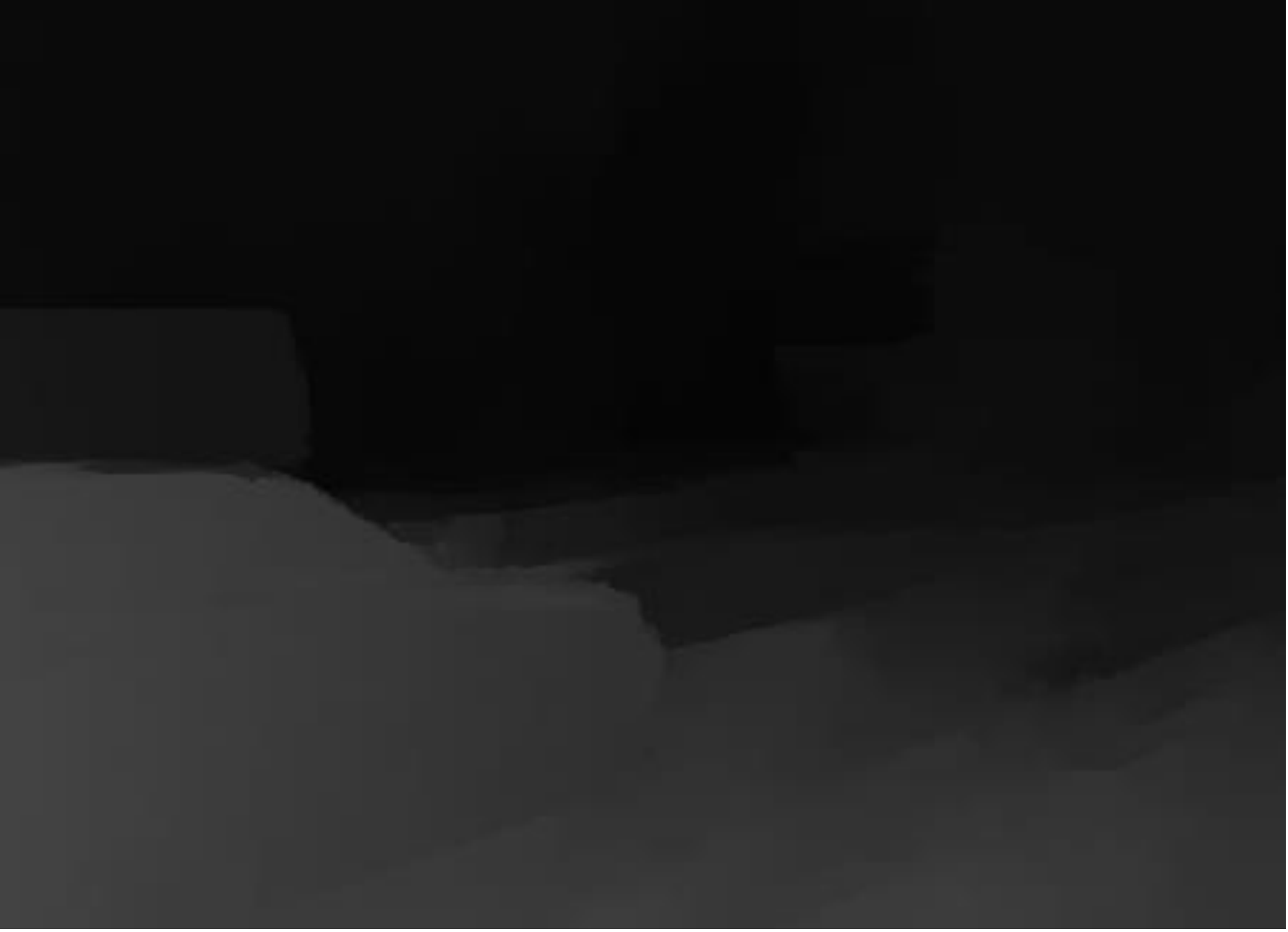}\\
\multicolumn{2}{c}{(c) Ferstl \etal ~\cite{ferstl2013image}} & \multicolumn{2}{c}{(d) Park \etal ~\cite{park2014high} } \\
\hspace{0.0cm}
\includegraphics[width=0.243\textwidth]{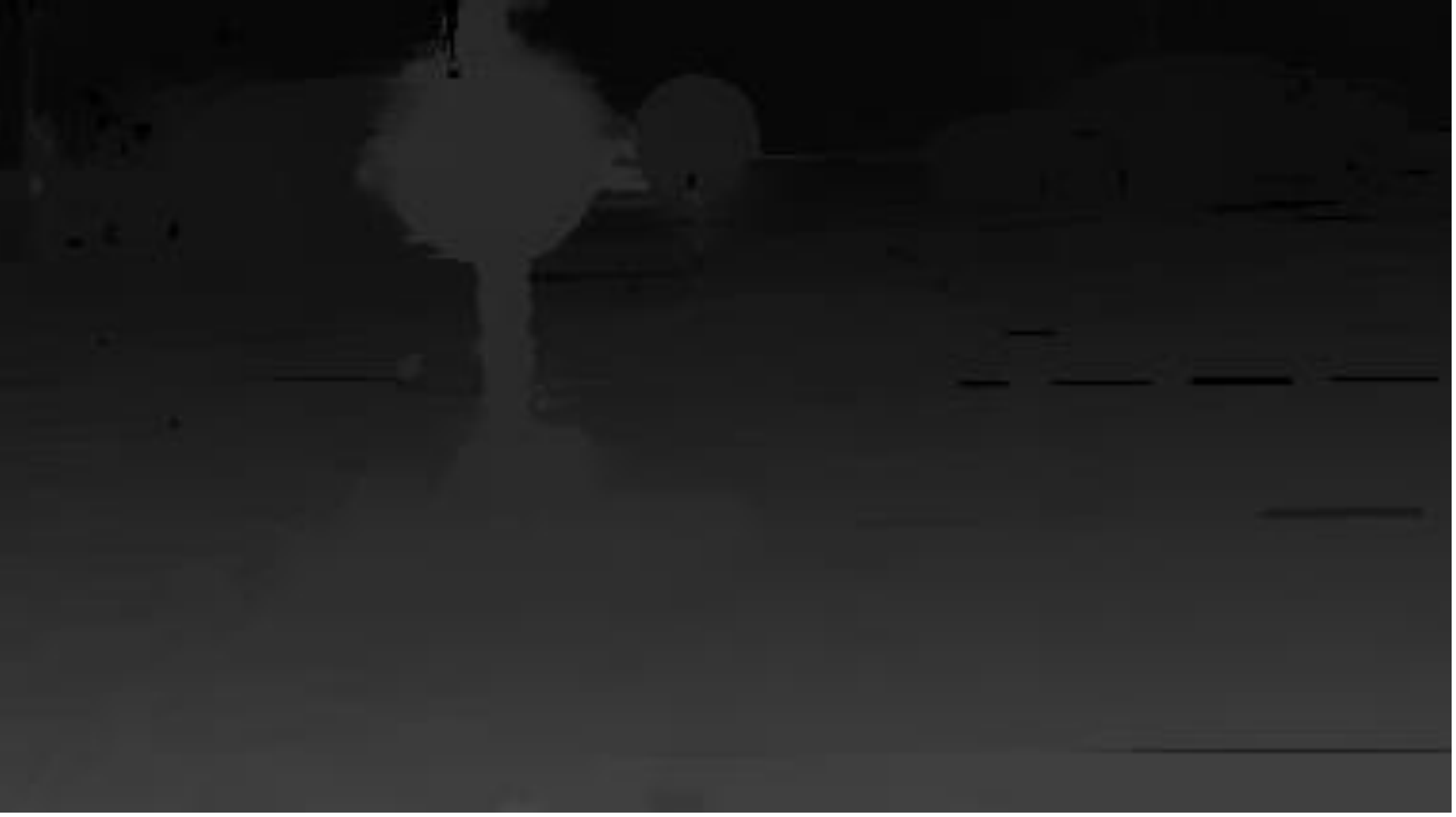}
&\hspace{0.0cm}
\includegraphics[width=0.189\textwidth]{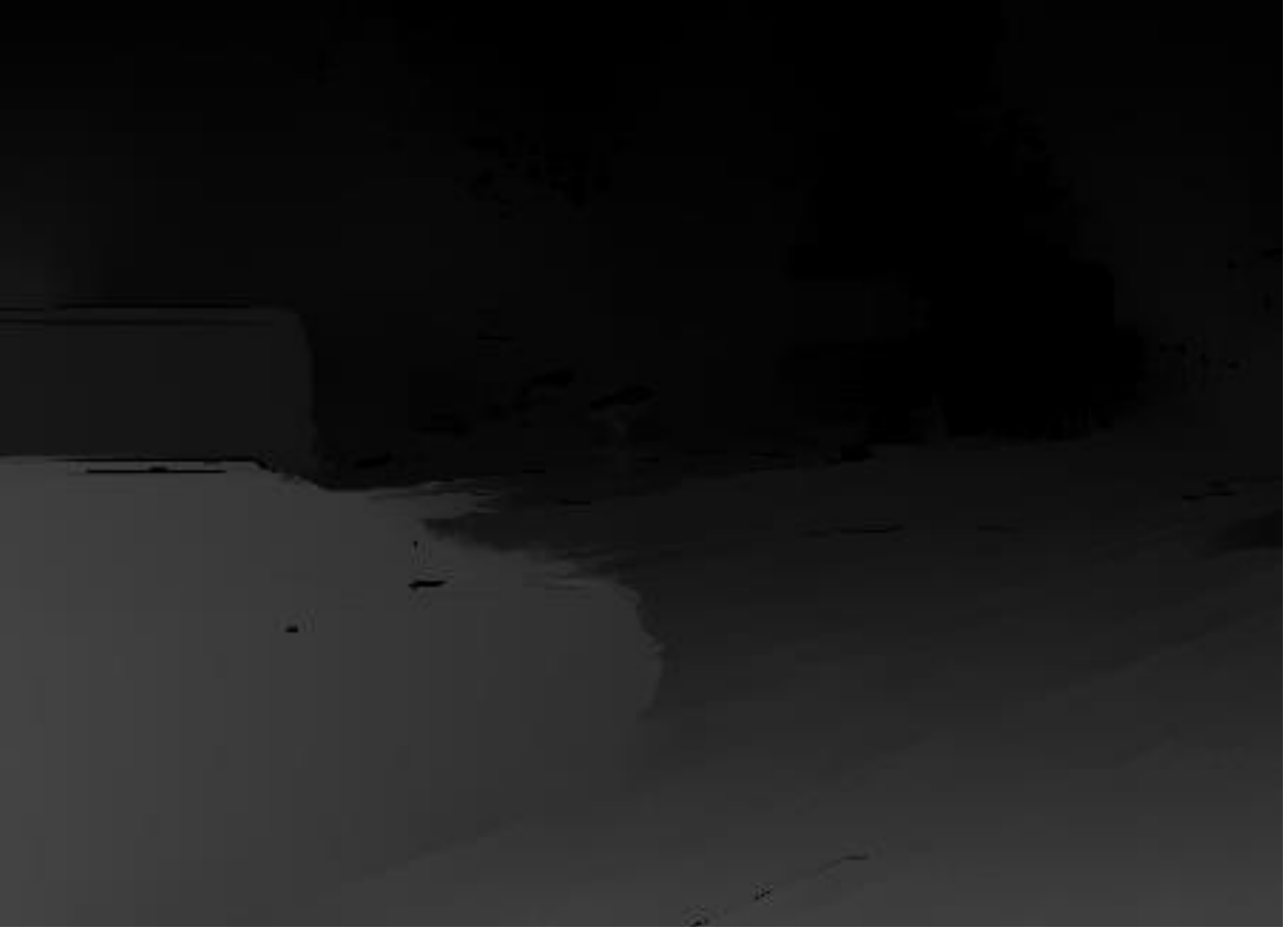}
&\hspace{0.0cm}
\includegraphics[width=0.243\textwidth]{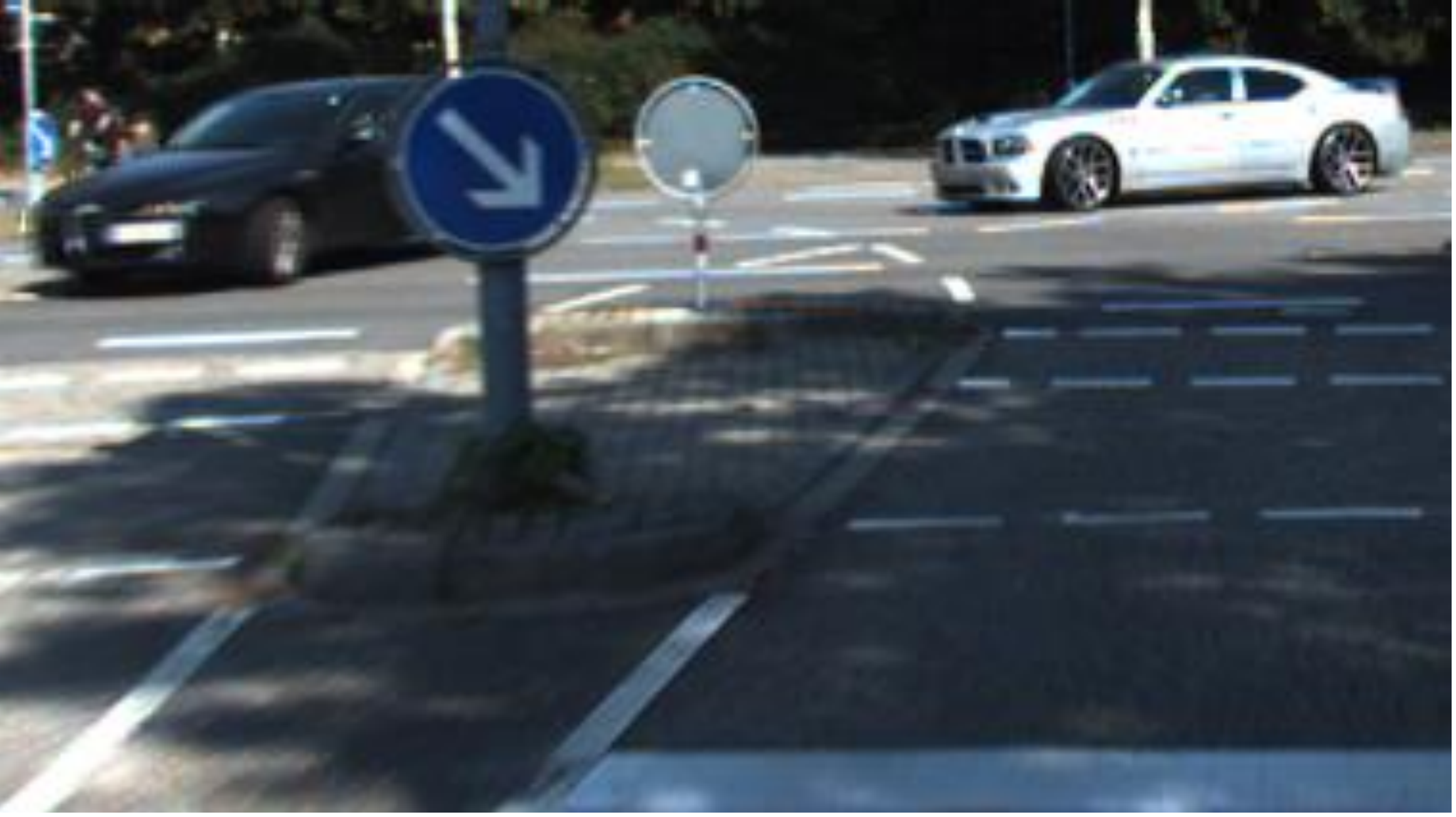}
&\hspace{0.0cm}
\includegraphics[width=0.189\textwidth]{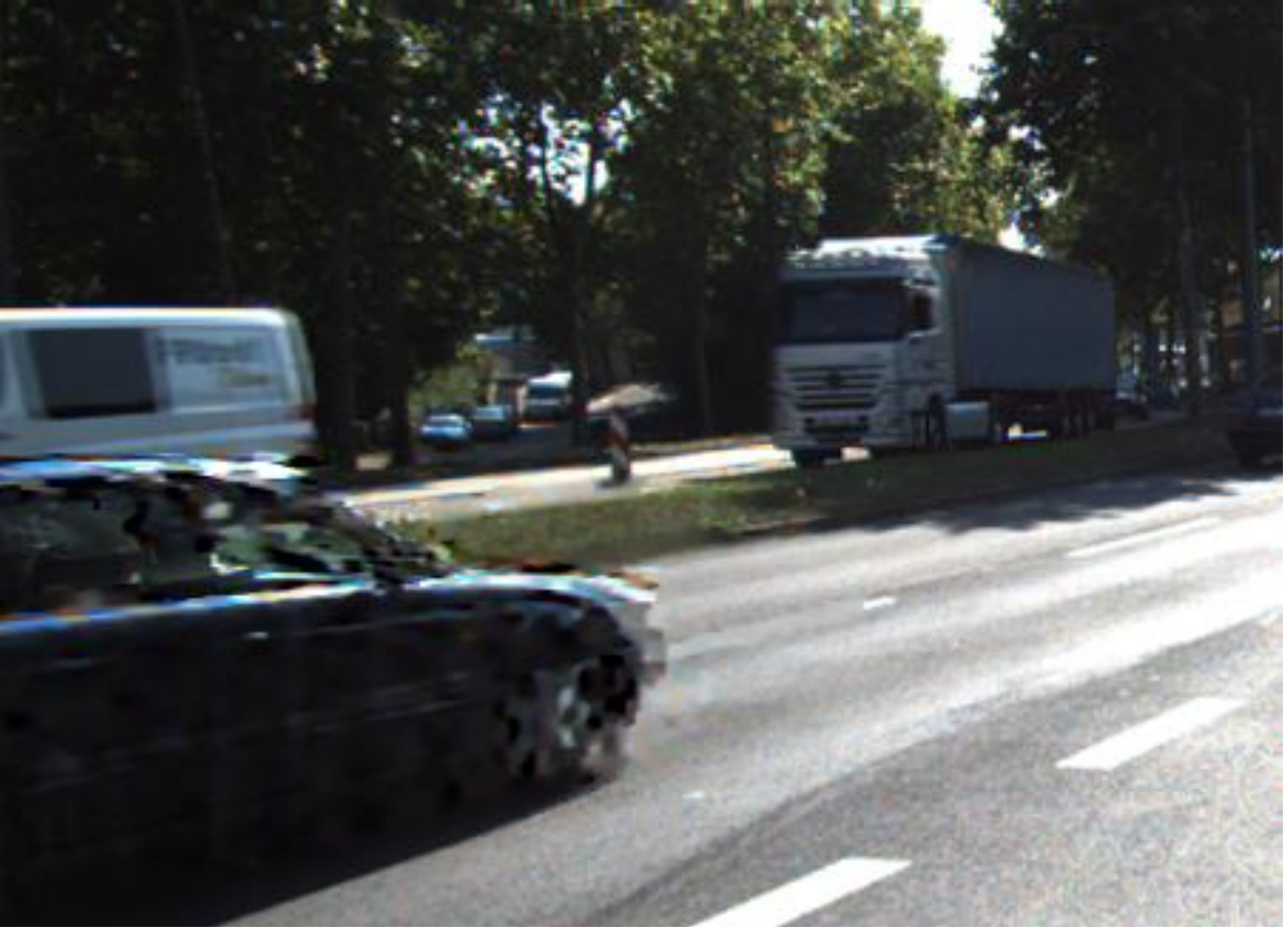}\\
\multicolumn{2}{c}{(e) Yang \etal ~\cite{yang2014color}} &\multicolumn{2}{c}{(f) Sellent \etal ~\cite{sellent2016stereo}} \\
\hspace{0.0cm}
\includegraphics[width=0.243\textwidth]{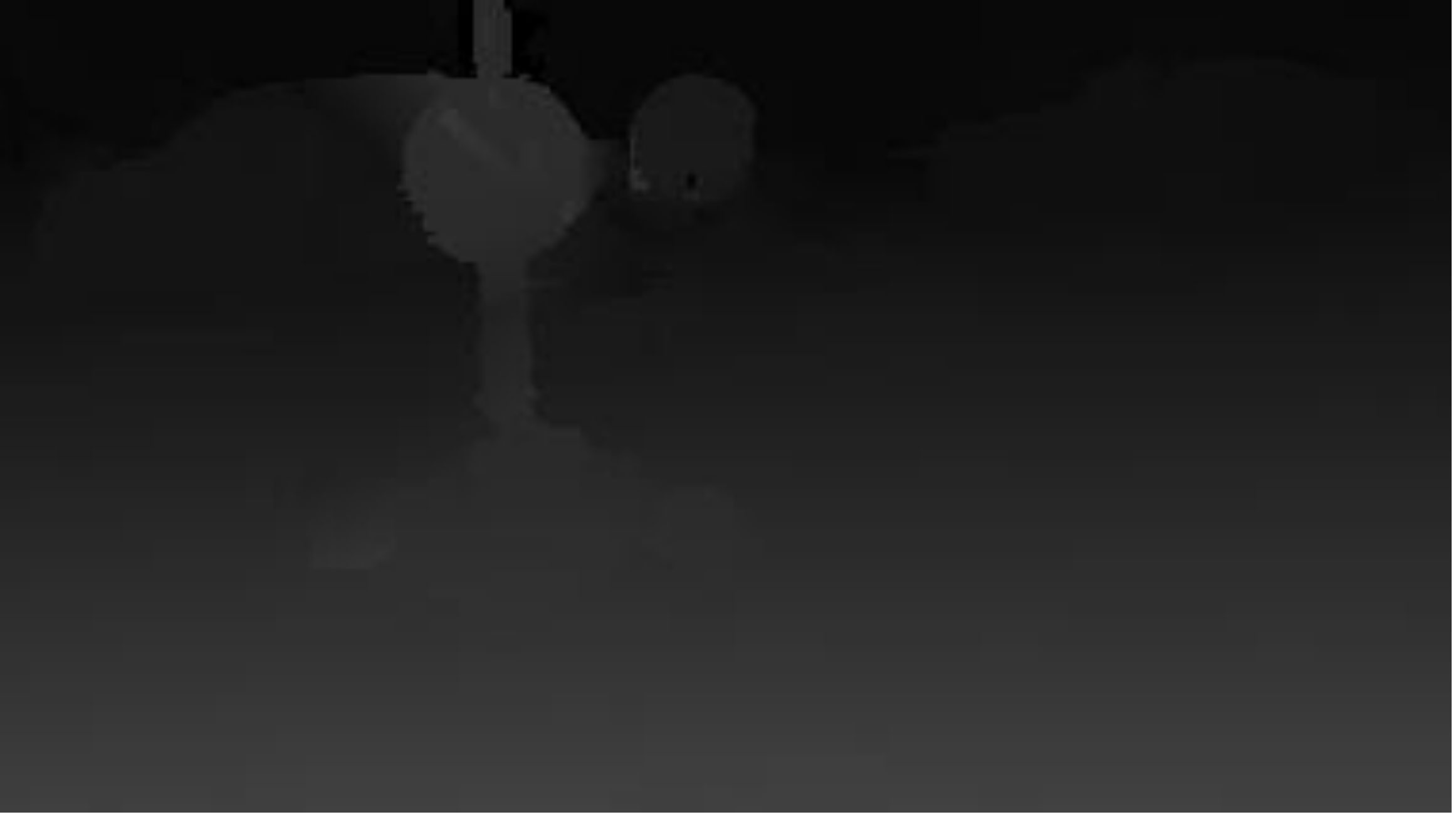}
&\hspace{0.0cm}
\includegraphics[width=0.189\textwidth]{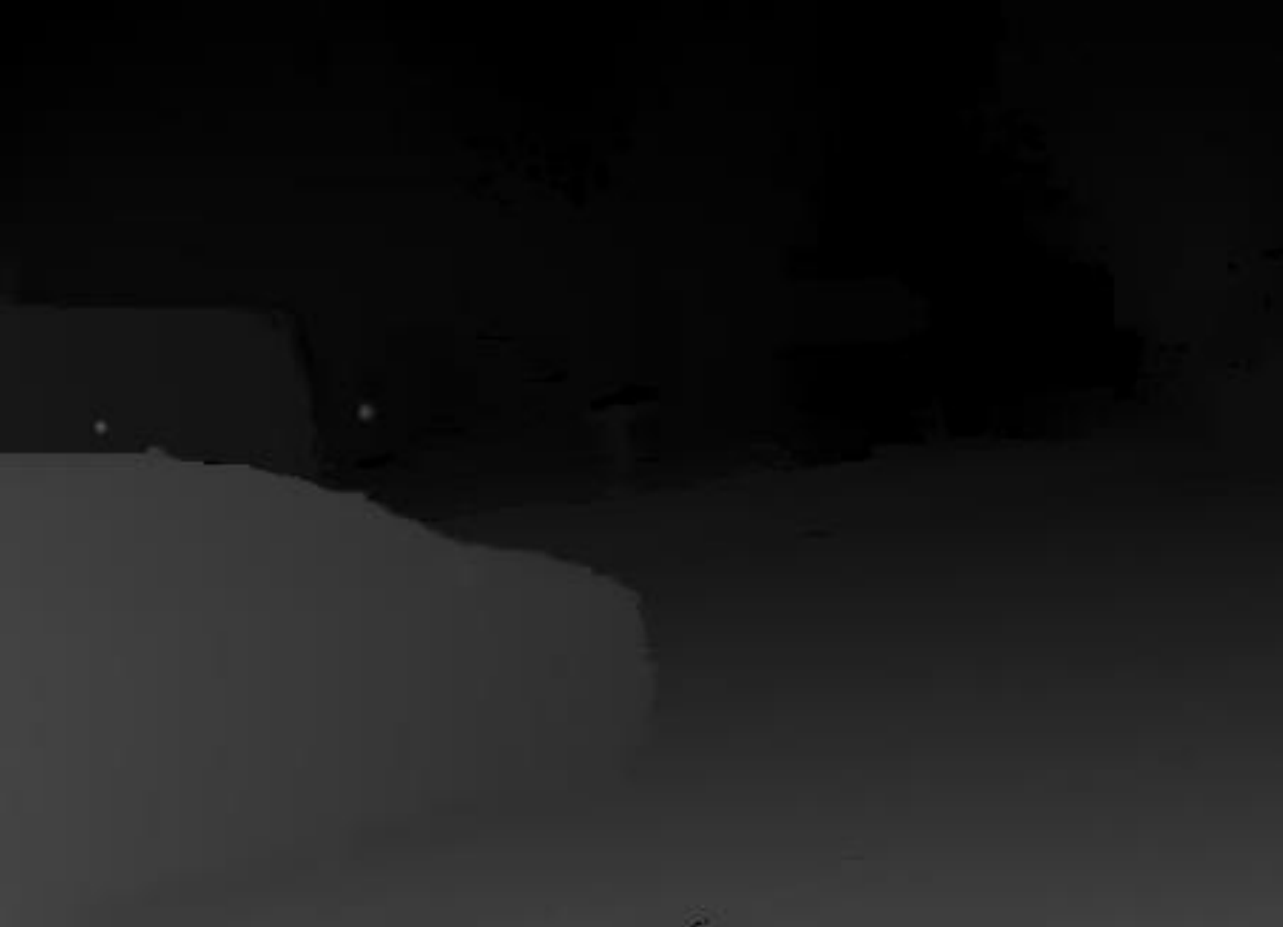}
&\hspace{0.0cm}
\includegraphics[width=0.243\textwidth]{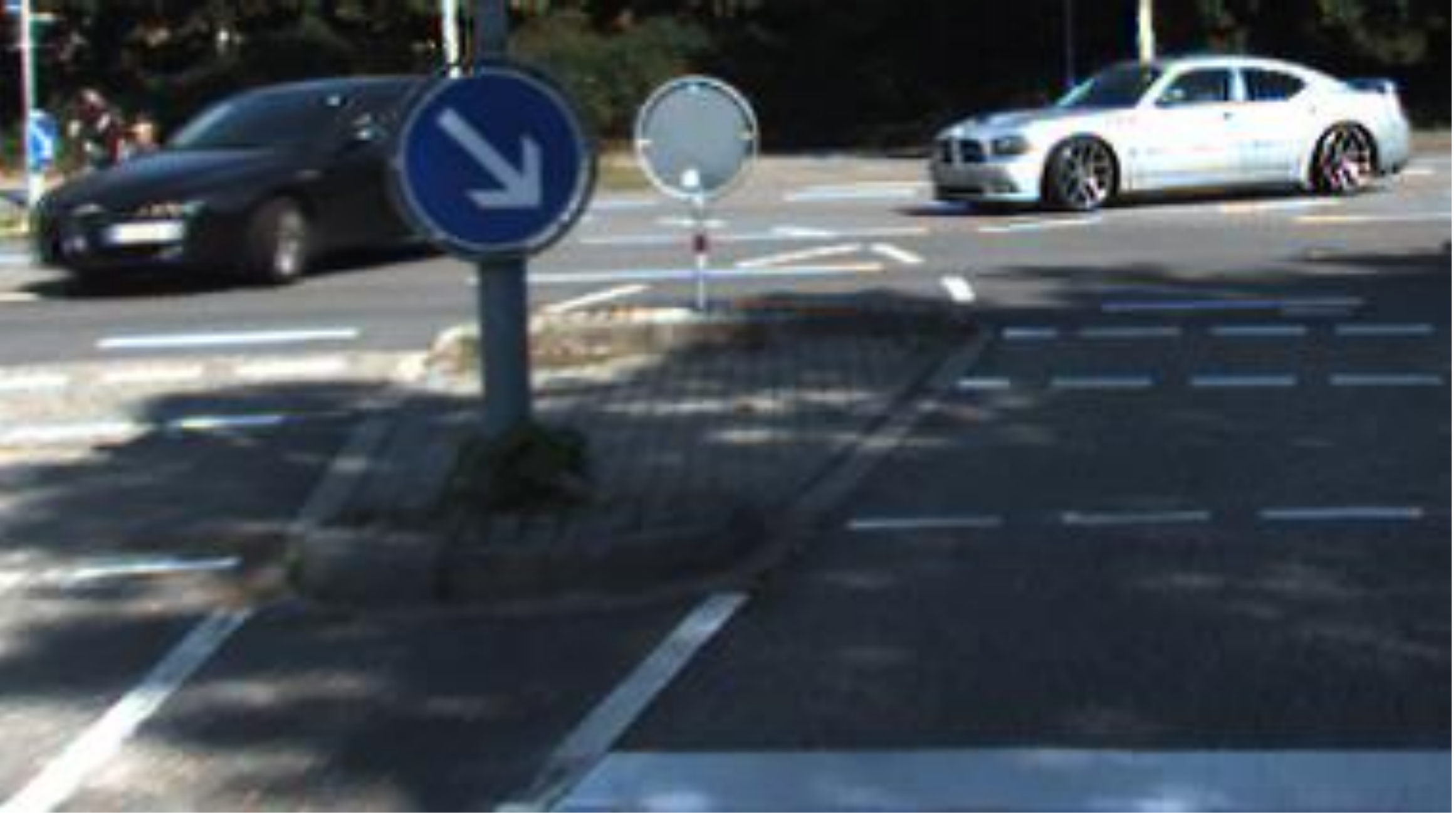}
&\hspace{0.0cm}
\includegraphics[width=0.189\textwidth]{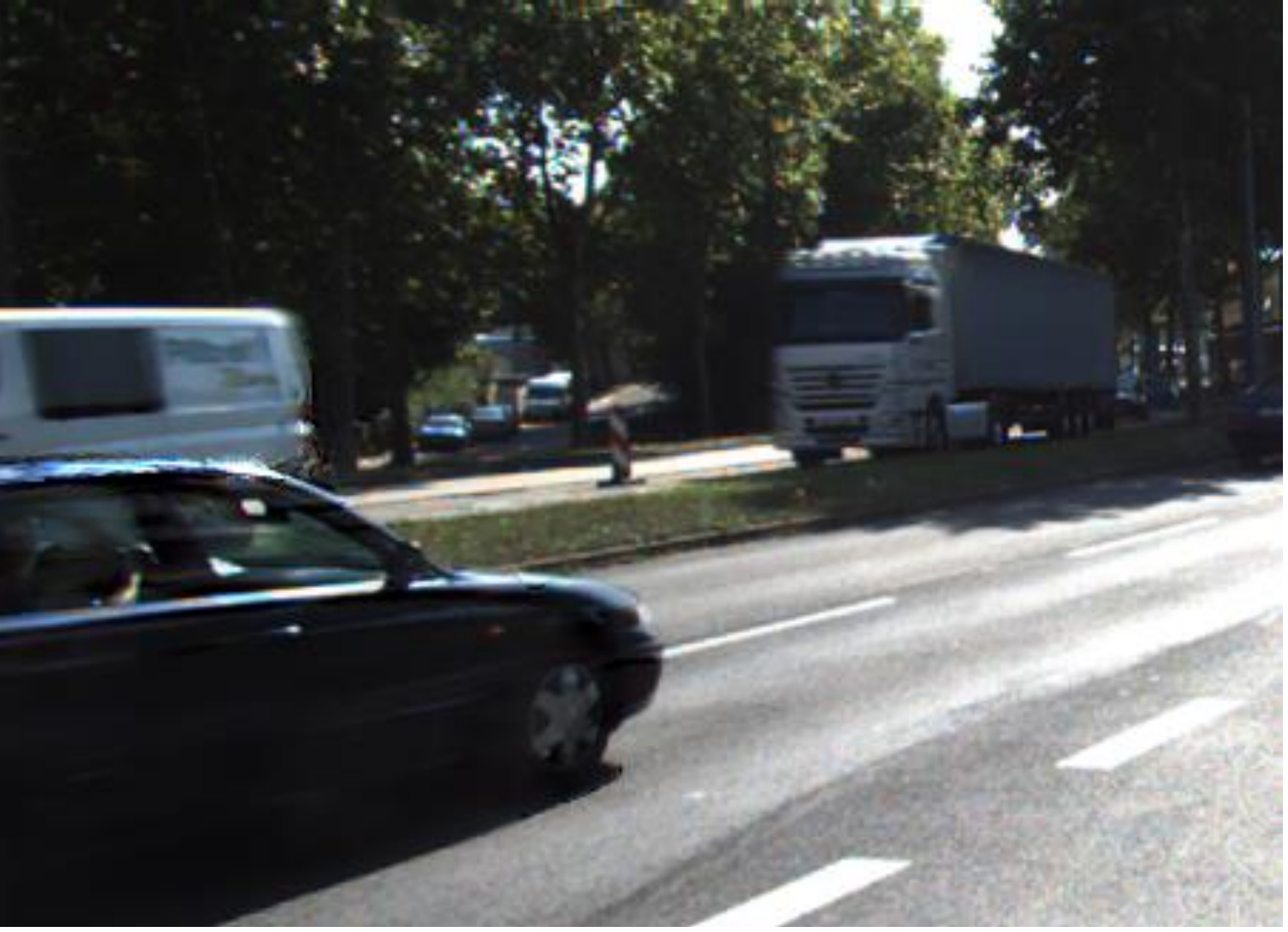}\\
\multicolumn{4}{c}{(g) Our completion depth map and deblurred image}  \\
\end{tabular}
\end{center}
\caption{
{Depth completion and image deblurring results on the KITTI dataset. (a) Input: sparse depth map. (b) Corresponding blurry color image (with ground-truth depth map in the corner). (c) Estimated depth map by \cite{ferstl2013image}. (d) Estimated depth by map \cite{park2014high}. (e) Estimated depth map by \cite{yang2014color}. (f) Deblurring result of \cite{sellent2016stereo}. (g) Our depth completion and deblurring result. Compared to the recent stereo deblurring method (i.e. \cite{sellent2016stereo}) and the remaining three state-of-the-art depth completion methods (i.e. \cite{ferstl2013image,park2014high, yang2014color}) shown above, our method achieves the best performance for both depth completion and deblurring. Best viewed on screen.}}
\label{fig:kitti_all}
\end{figure*}

\begin{figure*}
\begin{center}
\begin{tabular}{cccc}
\hspace{0.0cm}
\includegraphics[width=0.215\textwidth]{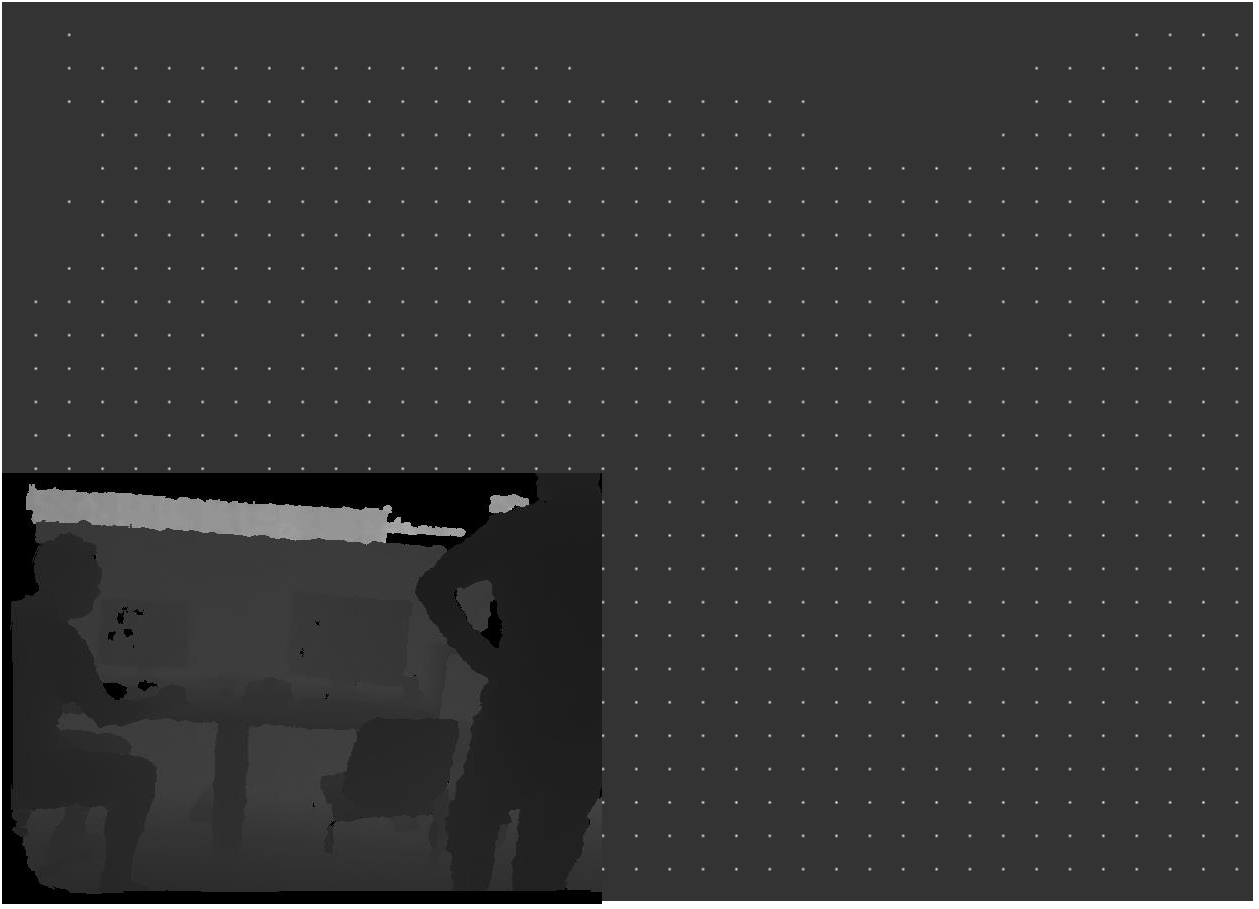}
&\hspace{0.0cm}
\includegraphics[width=0.203\textwidth]{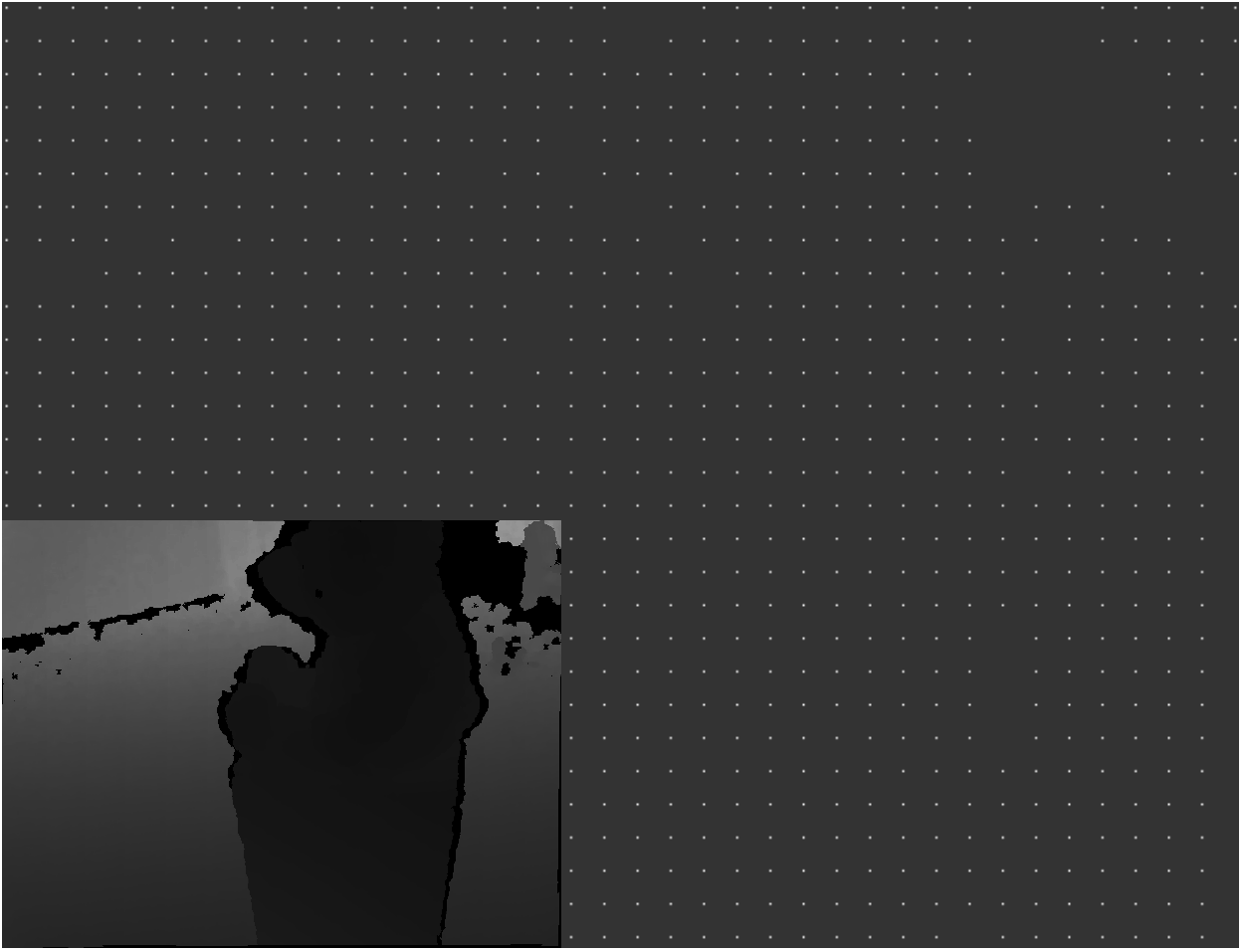}
&\hspace{0.0cm}
\includegraphics[width=0.215\textwidth]{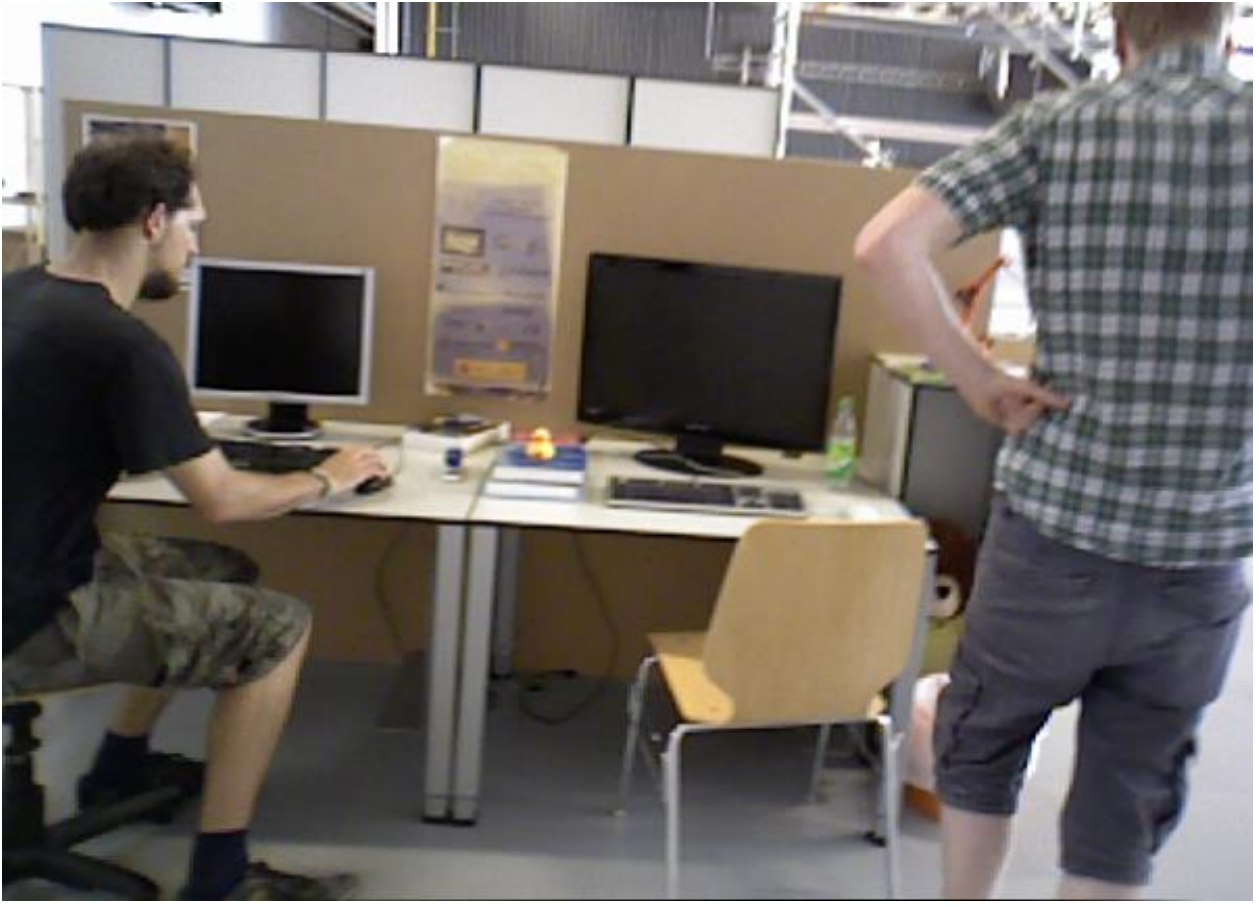}
&\hspace{0.0cm}
\includegraphics[width=0.203\textwidth]{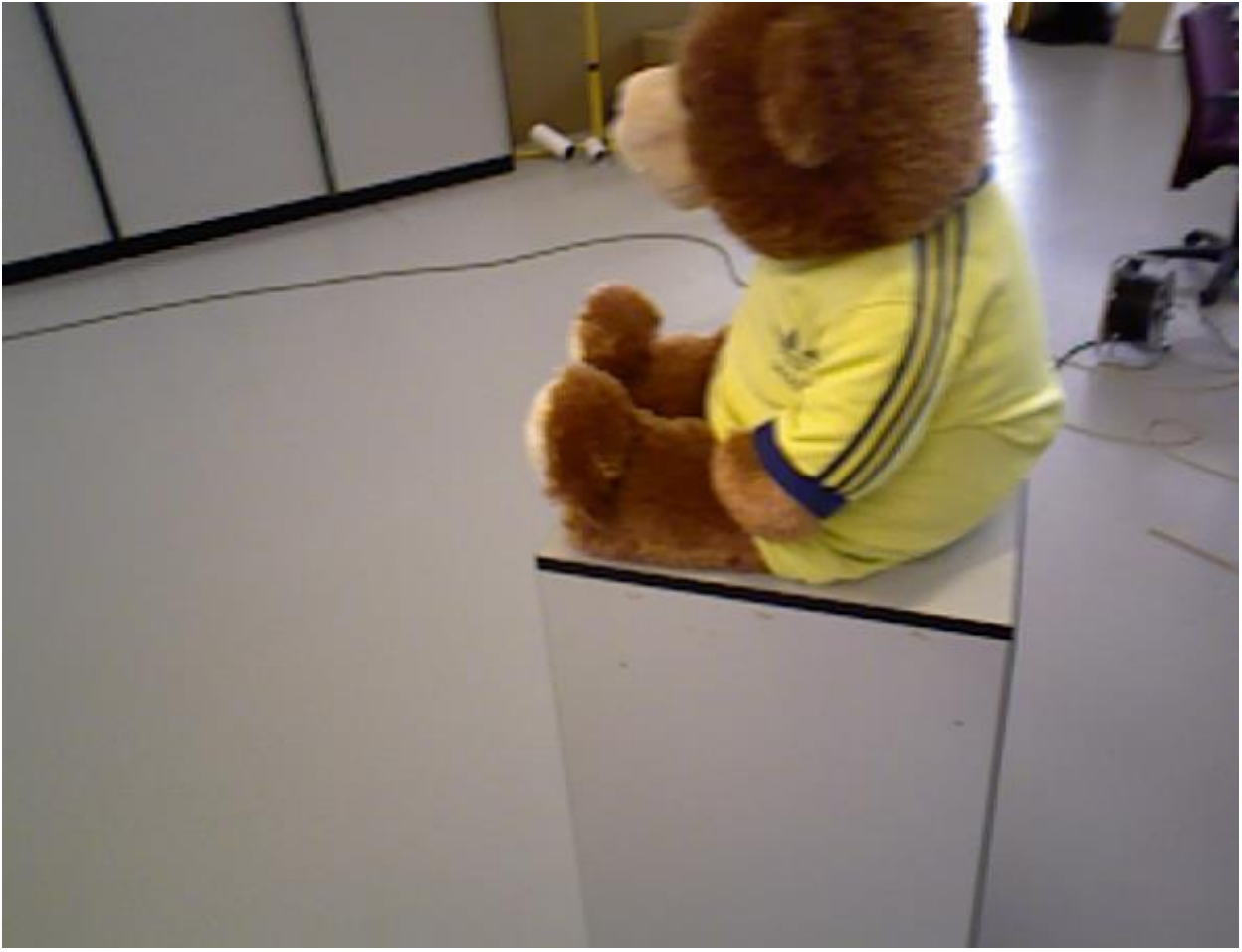}\\
\multicolumn{2}{c}{(a) Input incomplete and noisy depth map} &\multicolumn{2}{c}{(b) Input blurry image} \\
\hspace{0.0cm}
\includegraphics[width=0.215\textwidth]{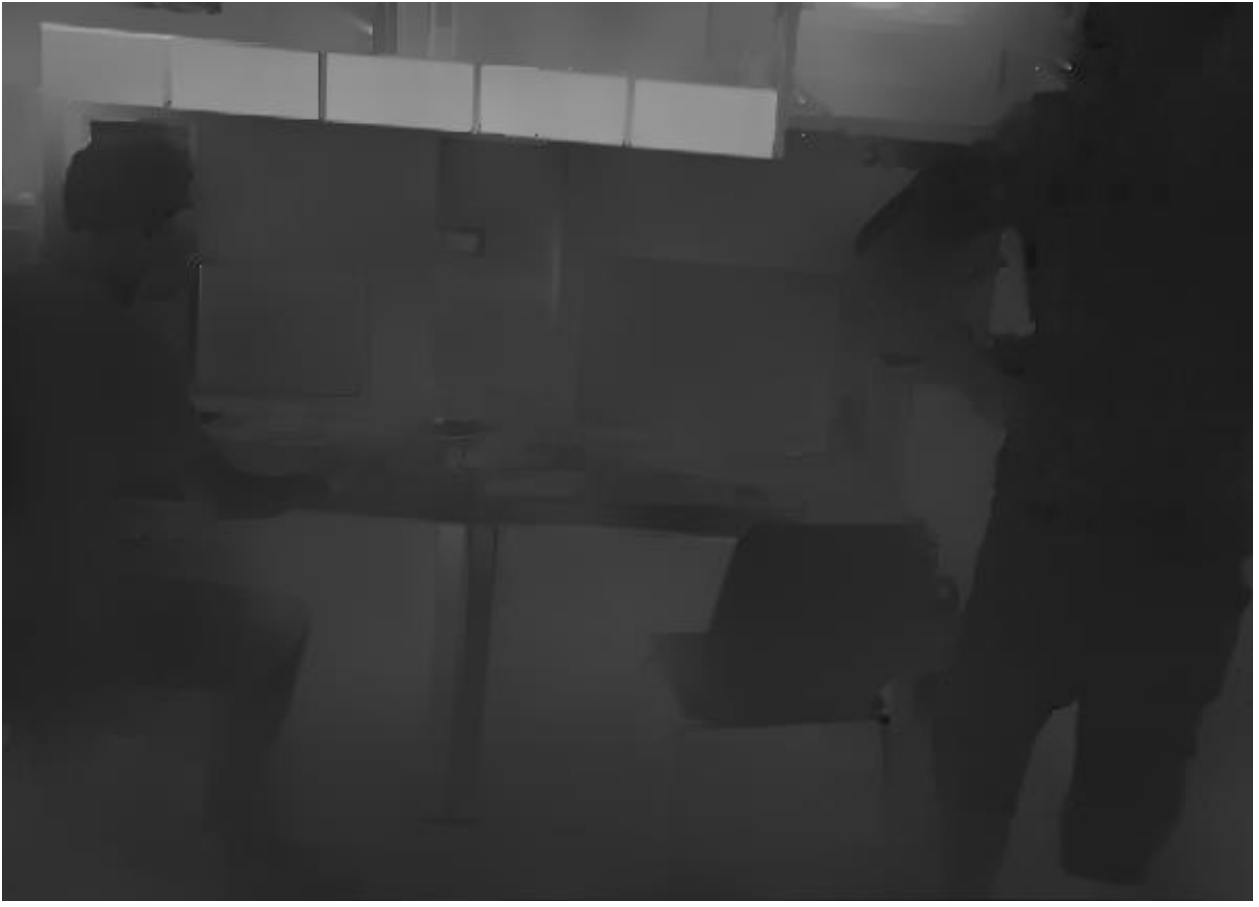}
&\hspace{0.0cm}
\includegraphics[width=0.203\textwidth]{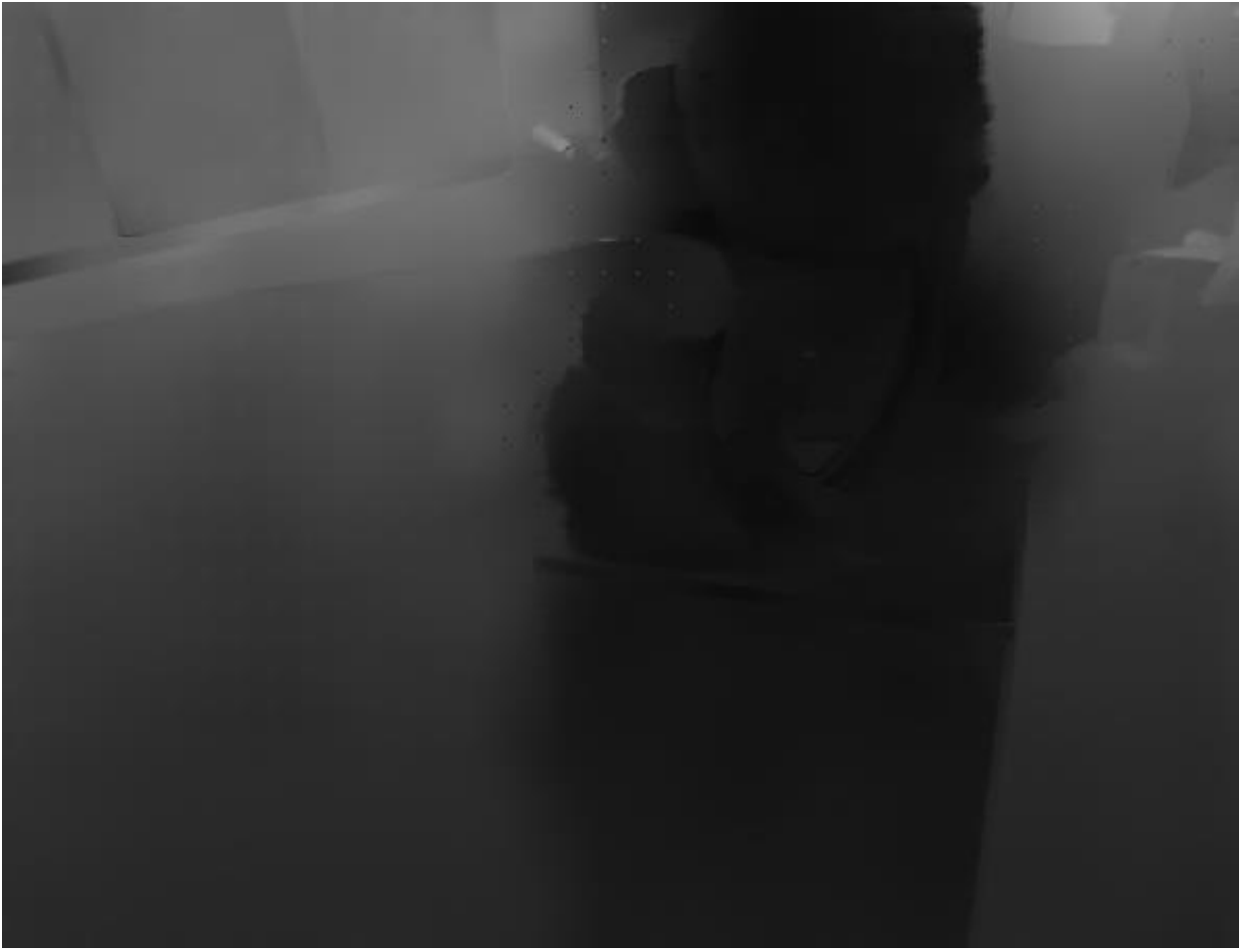}
&\hspace{0.0cm}
\includegraphics[width=0.215\textwidth]{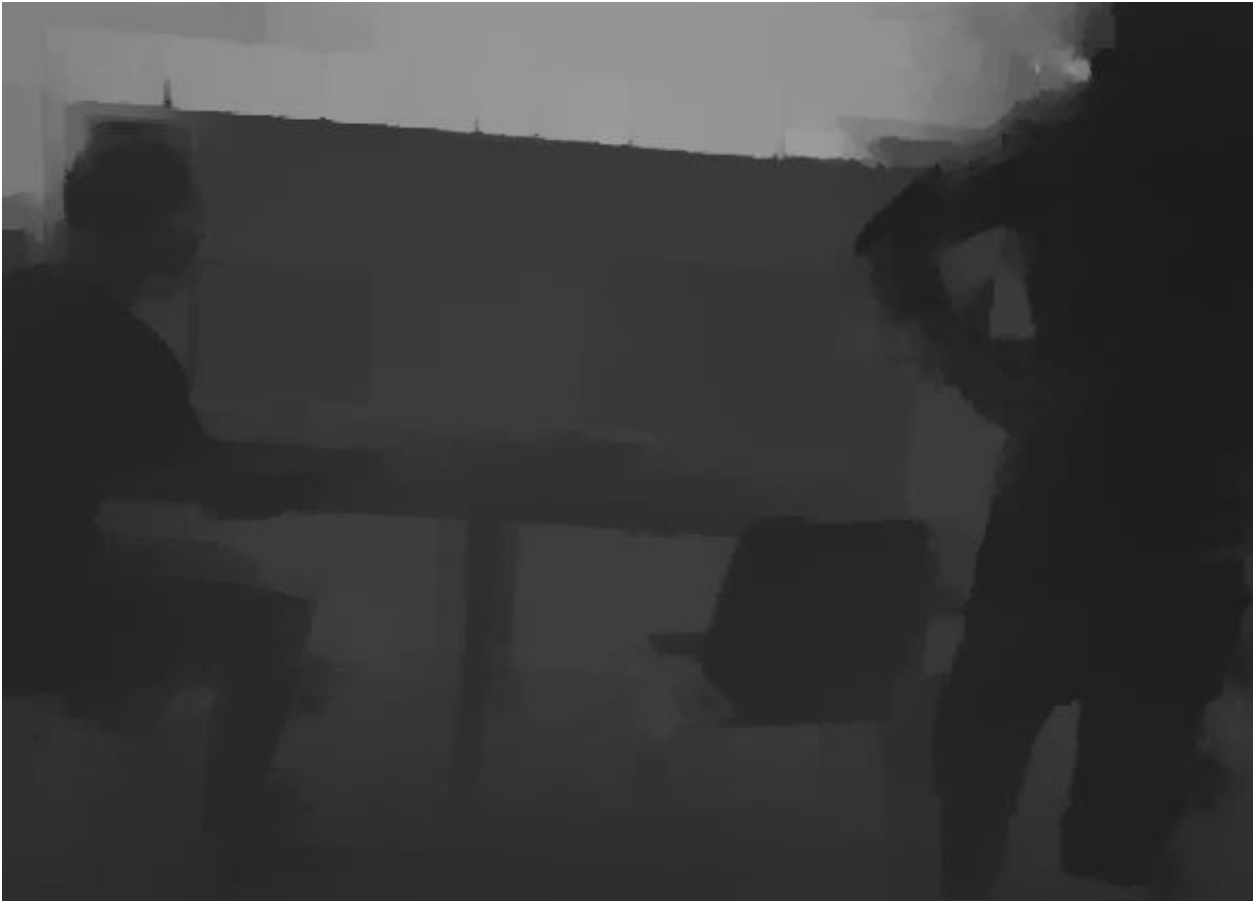}
&\hspace{0.0cm}
\includegraphics[width=0.203\textwidth]{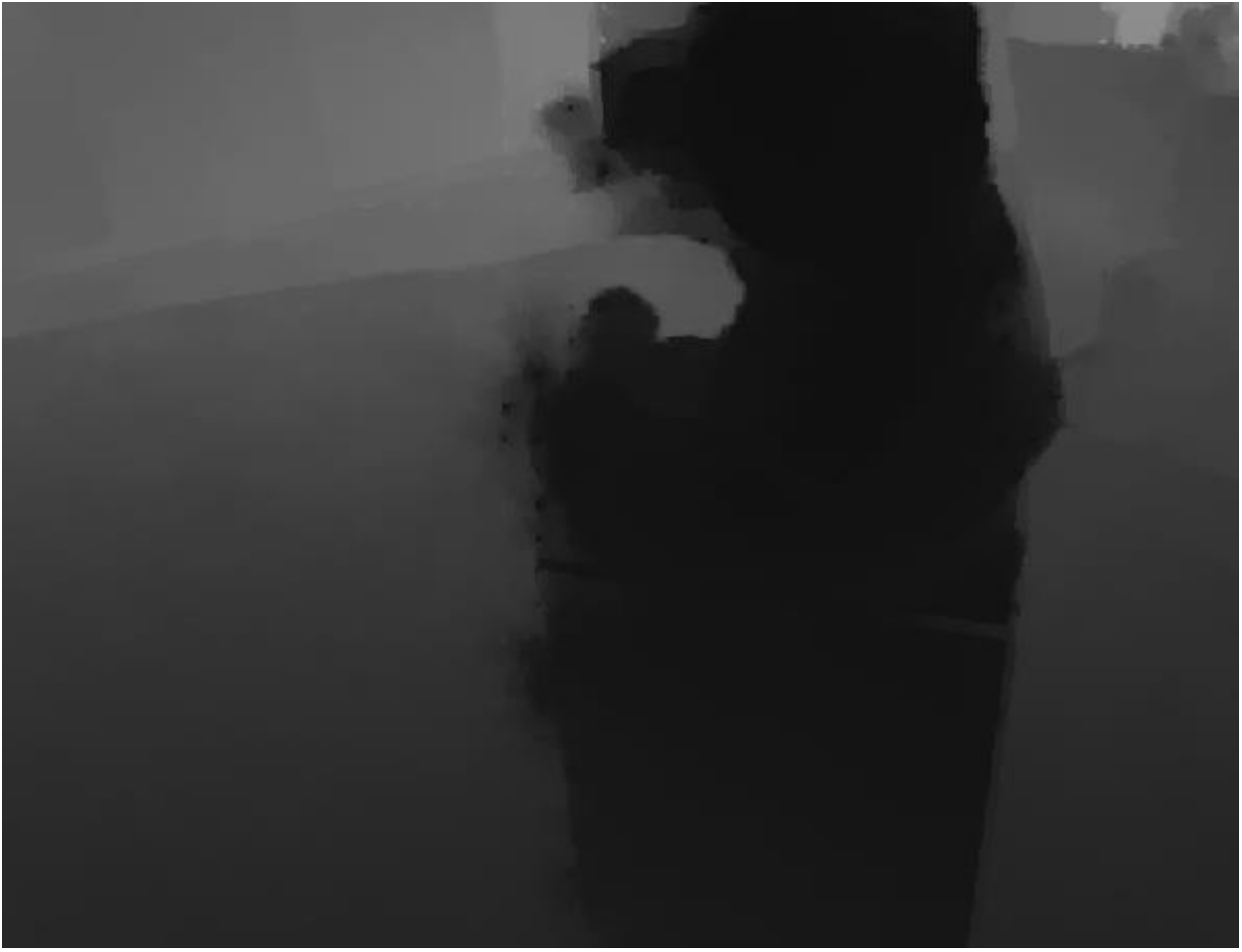}\\
\multicolumn{2}{c}{(c) Ferstl \etal ~\cite{ferstl2013image}} &\multicolumn{2}{c}{(d) Park \etal ~\cite{park2014high}} \\
\hspace{0.0cm}
\includegraphics[width=0.215\textwidth]{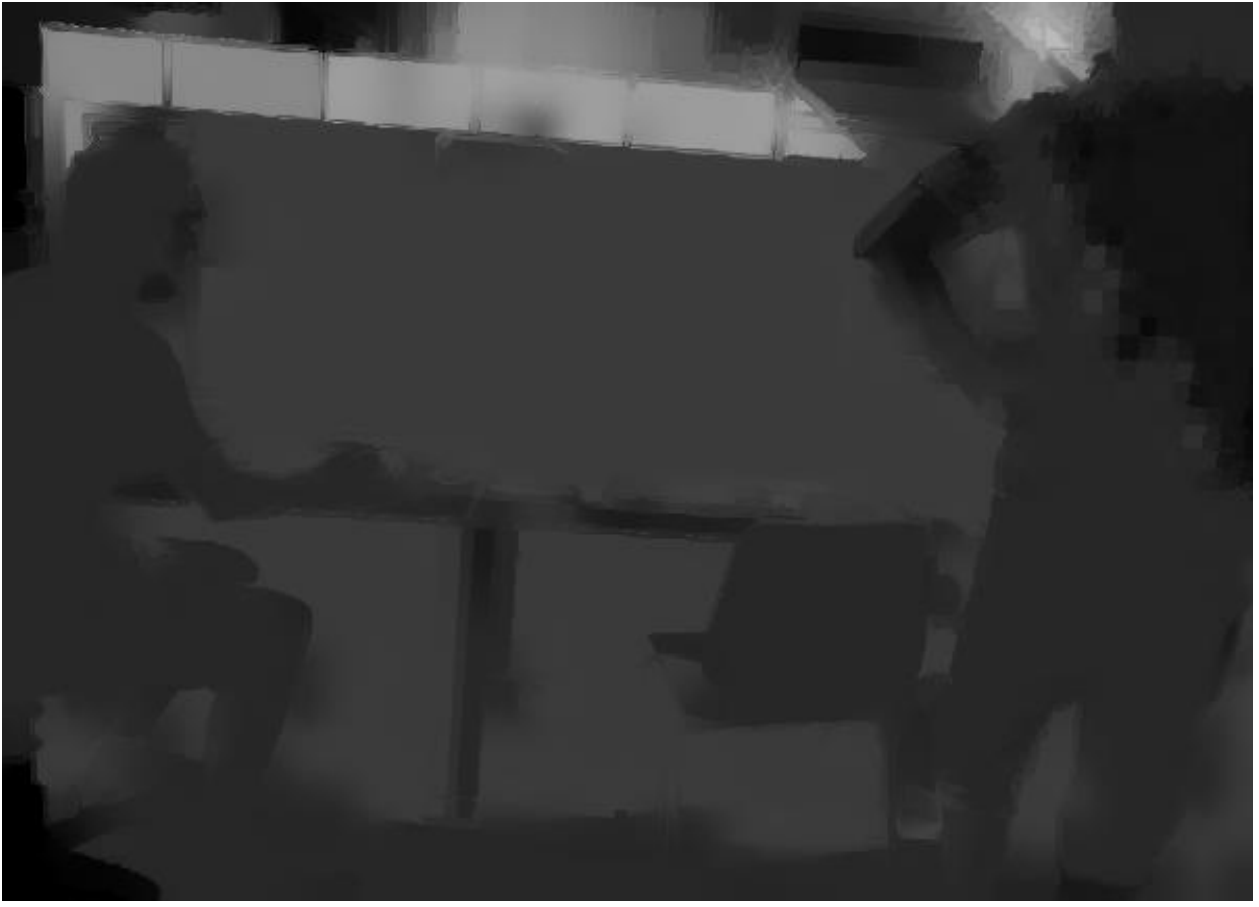}
&\hspace{0.0cm}
\includegraphics[width=0.203\textwidth]{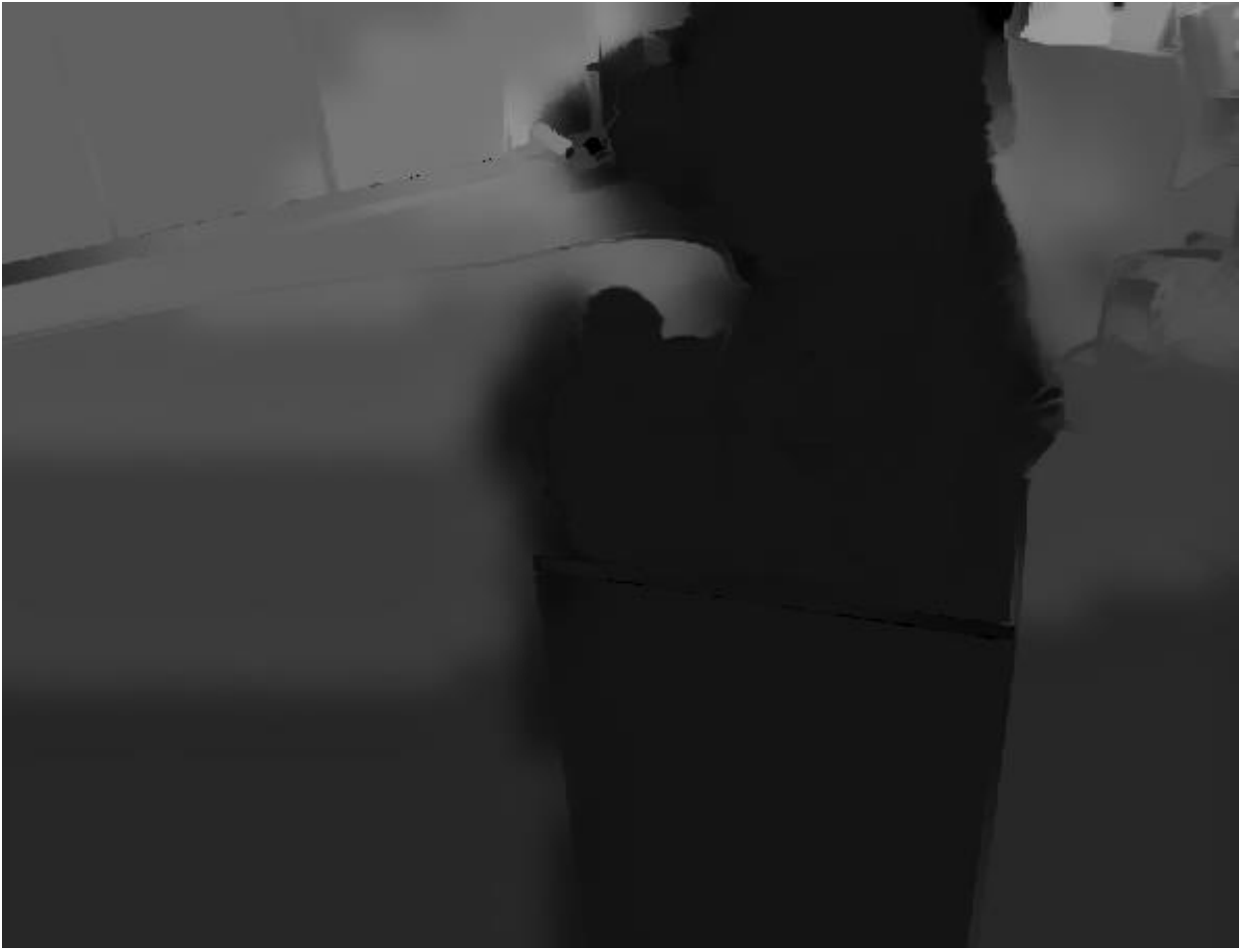}
&\hspace{0.0cm}
\includegraphics[width=0.215\textwidth]{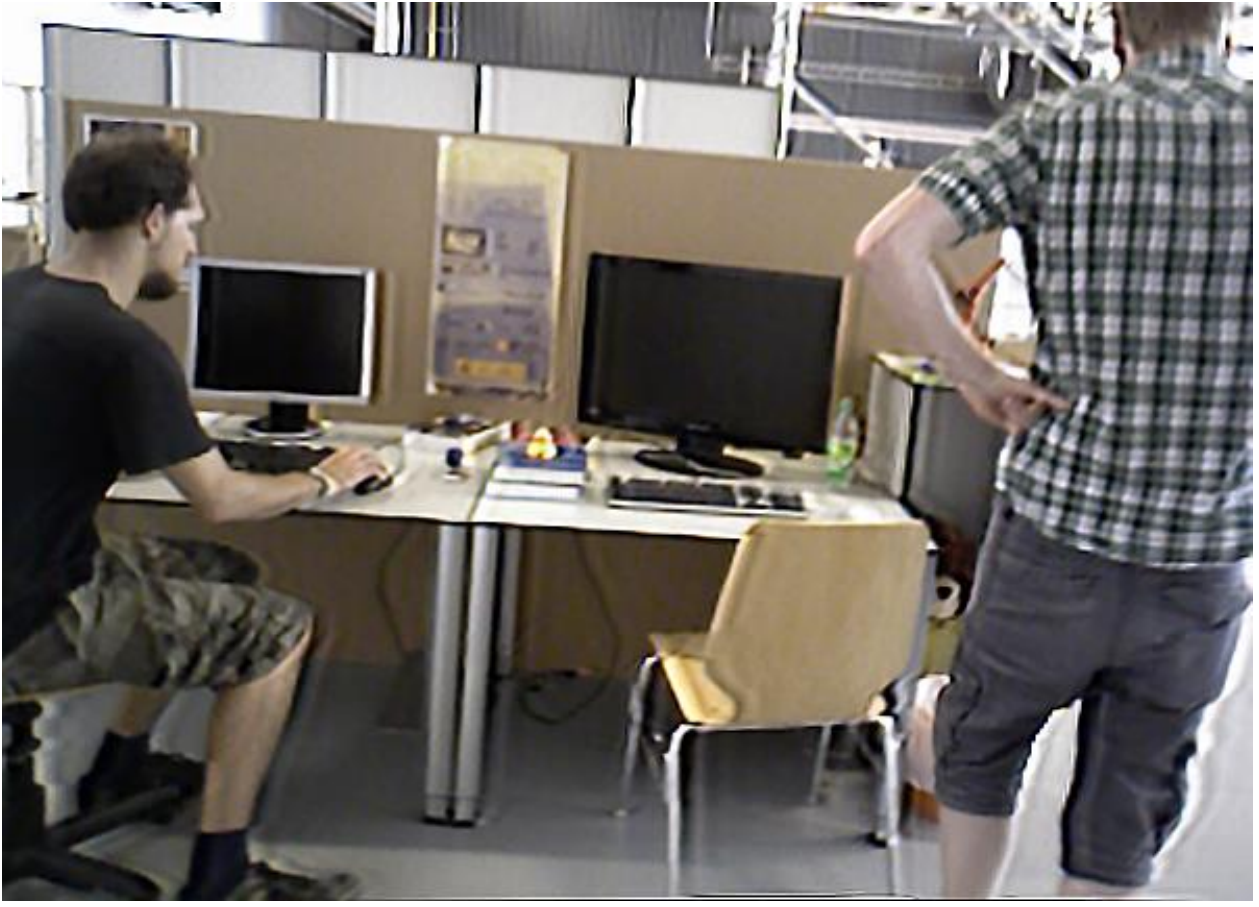}
&\hspace{0.0cm}
\includegraphics[width=0.203\textwidth]{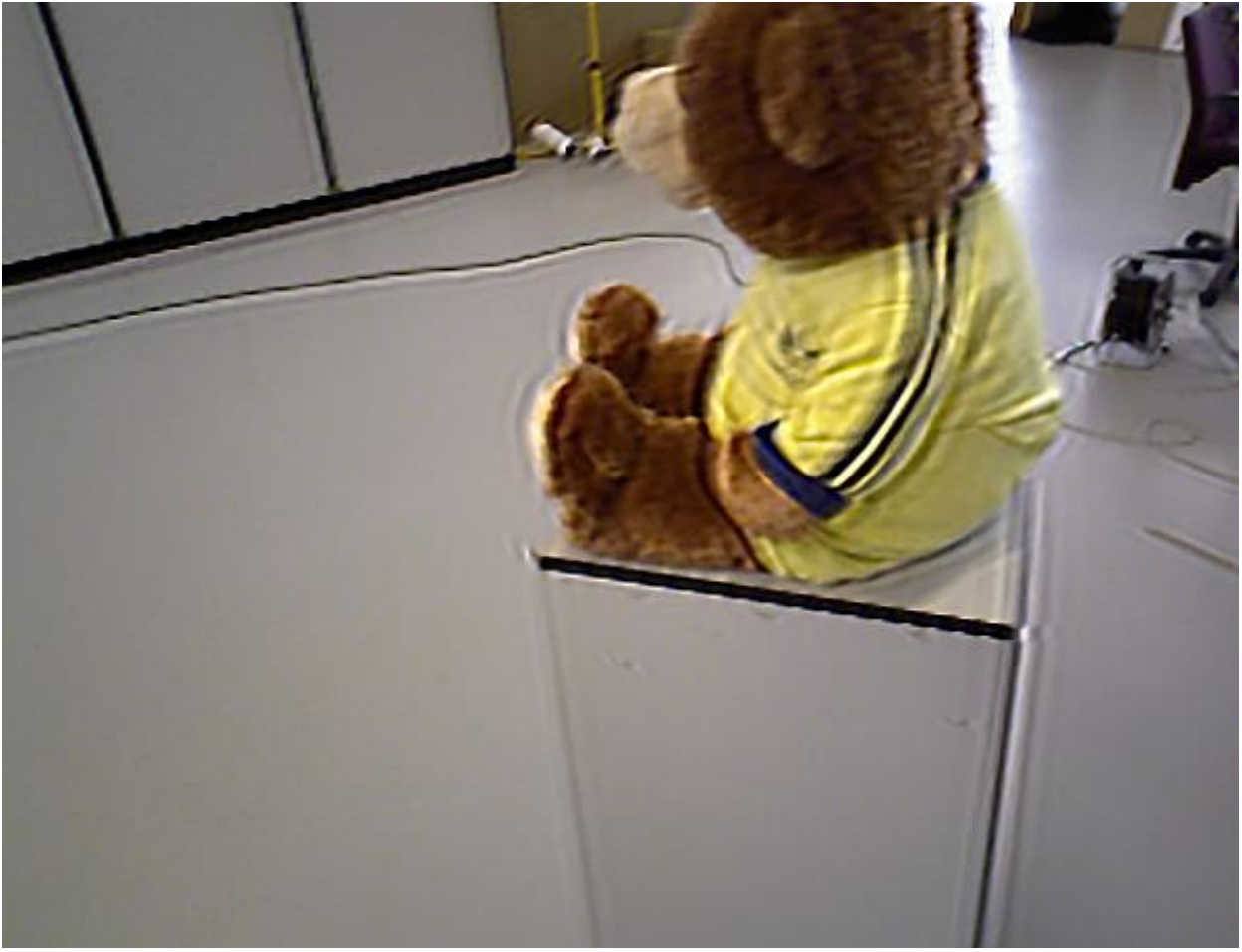}\\
\multicolumn{2}{c}{(e) Yang \etal ~\cite{yang2014color}} &\multicolumn{2}{c}{(f) Kim and Lee ~\cite{hyun2015generalized}}\\
\hspace{0.0cm}
\includegraphics[width=0.215\textwidth]{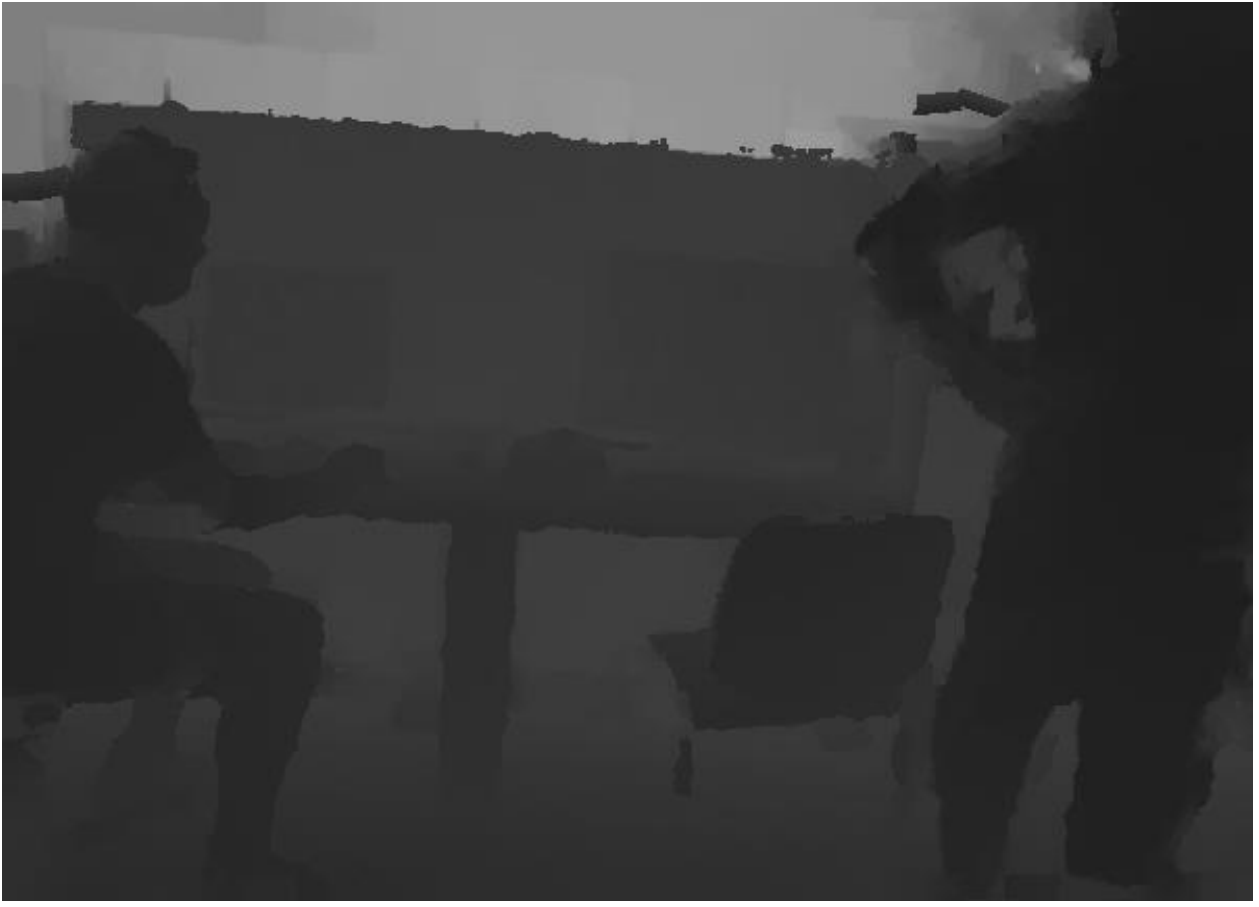}
&\hspace{0.0cm}
\includegraphics[width=0.203\textwidth]{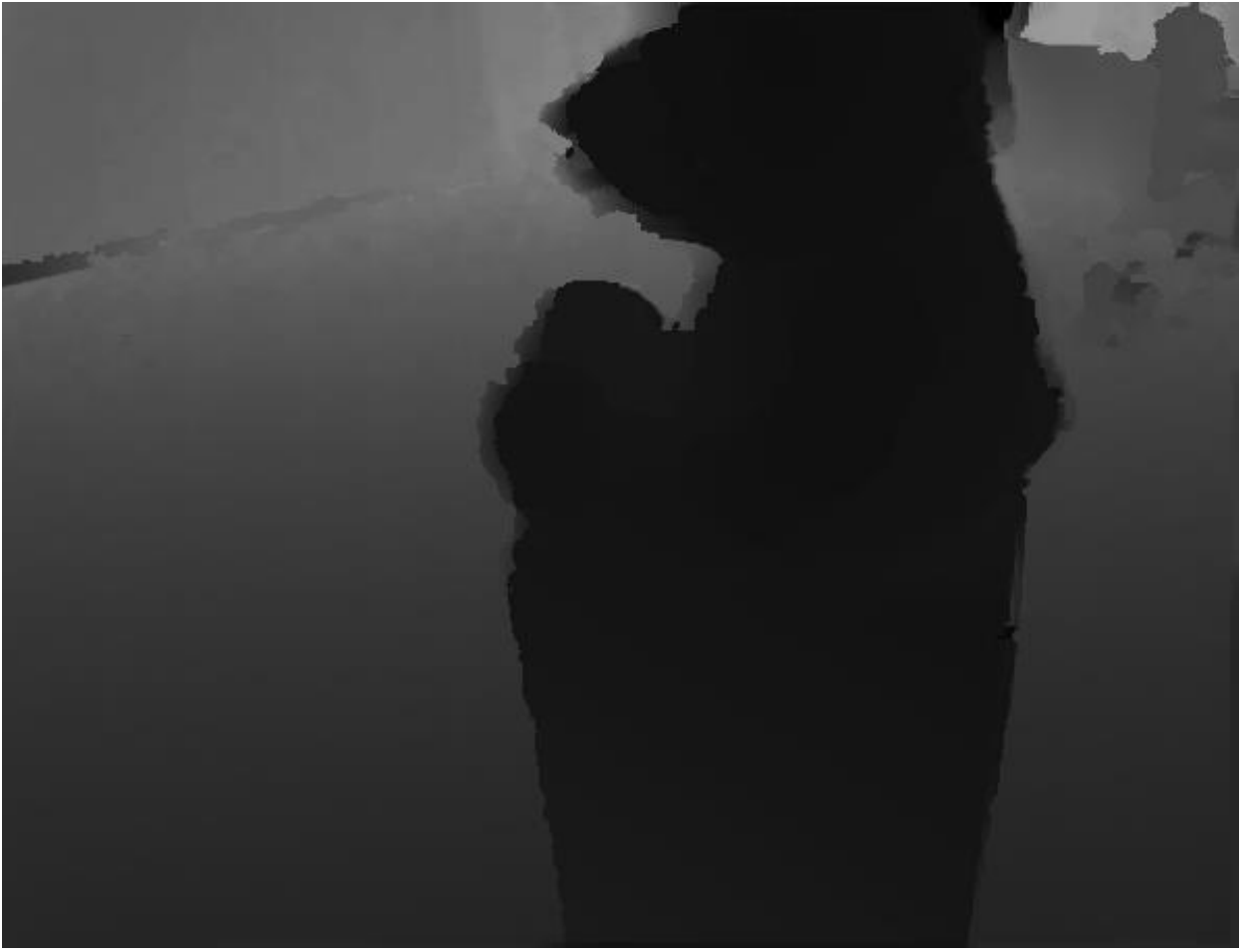}
&\hspace{0.0cm}
\includegraphics[width=0.215\textwidth]{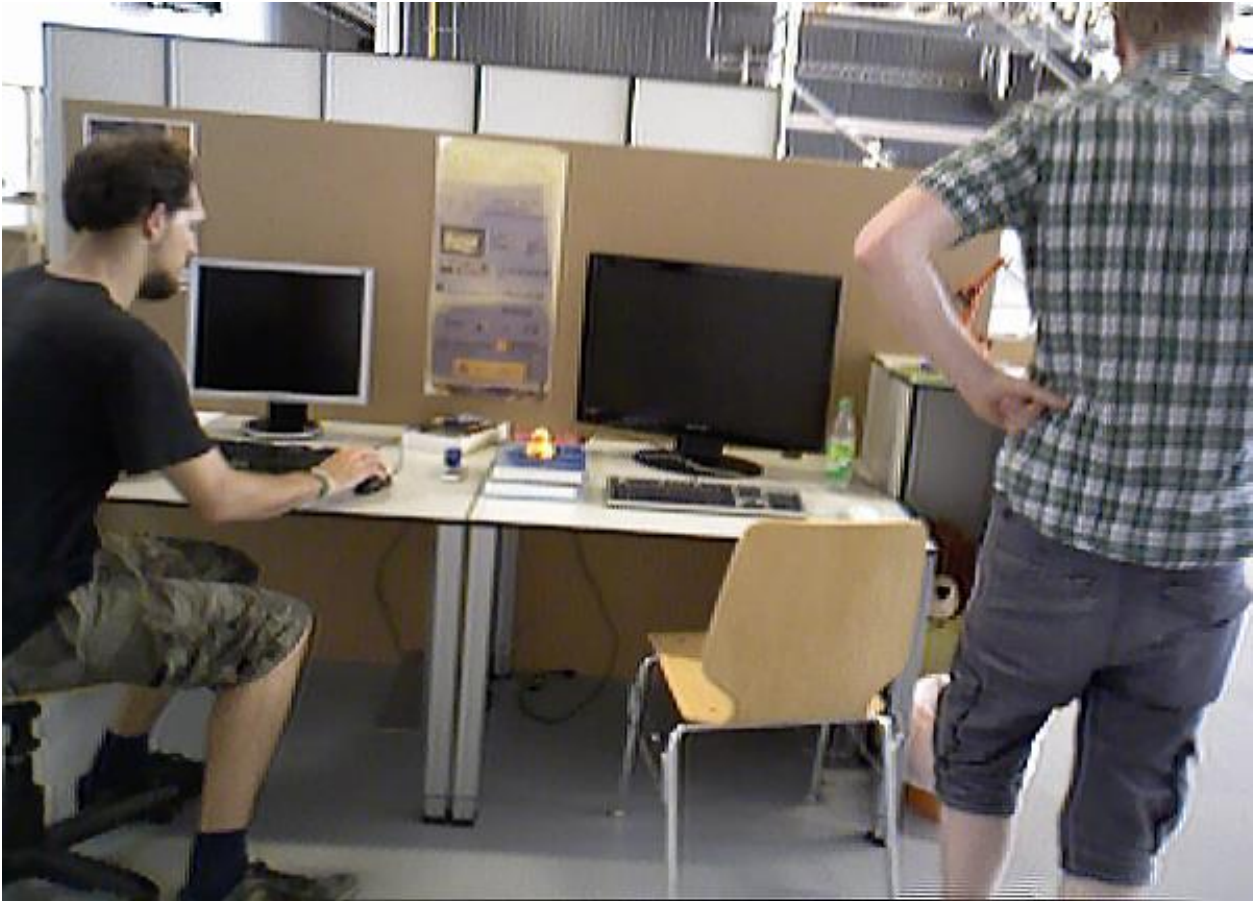}
&\hspace{0.0cm}
\includegraphics[width=0.203\textwidth]{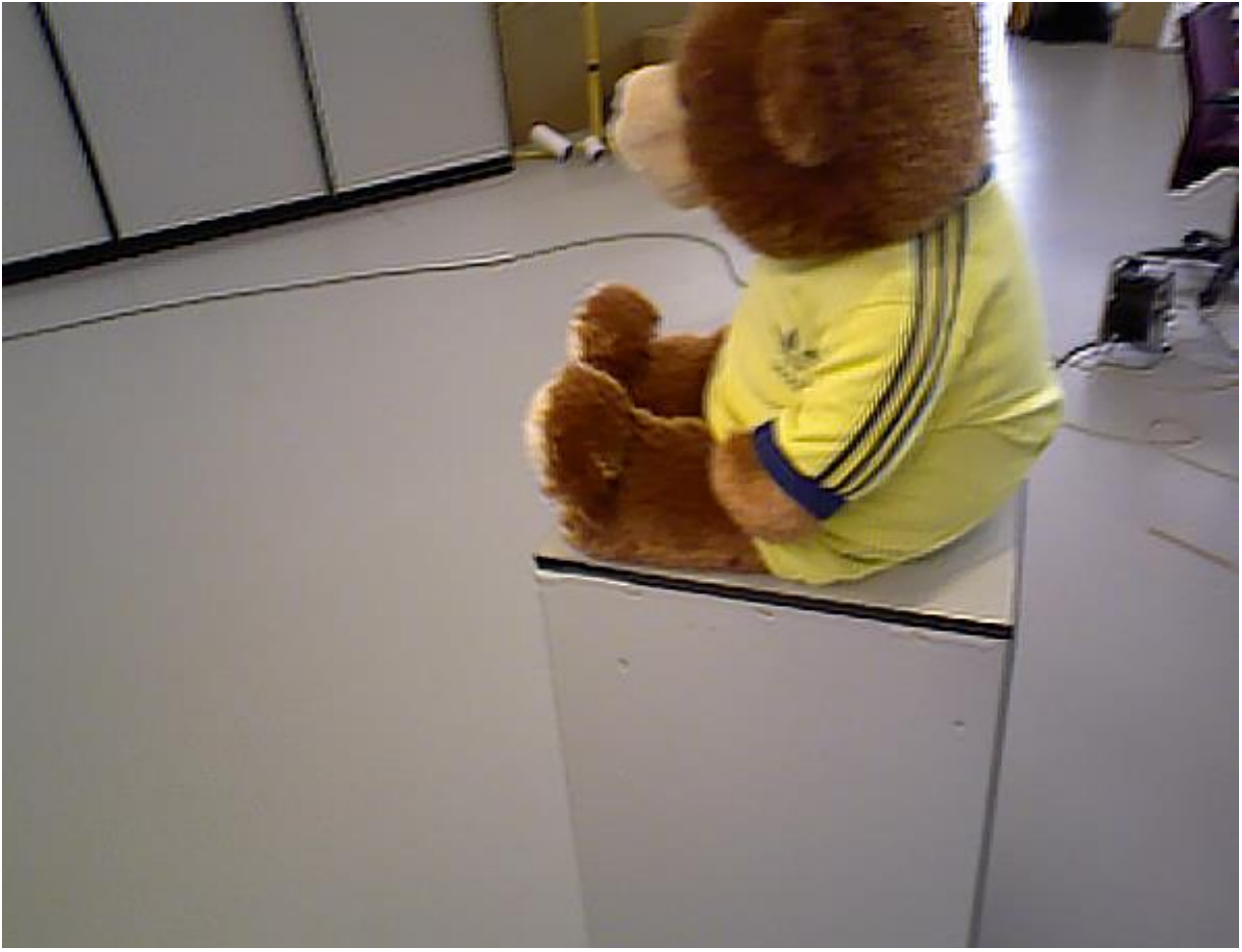}\\
\multicolumn{4}{c}{(g) Our completion depth map and deblurred image}\\
\end{tabular}
\end{center}
\caption{
{Depth completion and image deblurring results on the TUM dataset. (a) Input: incomplete and noisy depth map (with ground-truth depth map in the corner). (b) Corresponding blurry color image. (c) Estimated depth map by \cite{ferstl2013image}. (d) Estimated depth map by \cite{park2014high}. (e) Estimated depth map by \cite{yang2014color}. (f) Deblurring result of \cite{hyun2015generalized}. (g) Our depth completion and deblurring result. Compared to the monocular deblurring method (i.e. \cite{hyun2015generalized}) and the remaining three state-of-the-art depth completion methods (i.e. \cite{ferstl2013image,park2014high, yang2014color}) shown above, our method achieves the best performance for both depth completion and deblurring. Best viewed on screen.}
}
\label{fig:bear_all}
\end{figure*}
\begin{figure}[b]
\begin{center}
\includegraphics[width=0.225\textwidth]{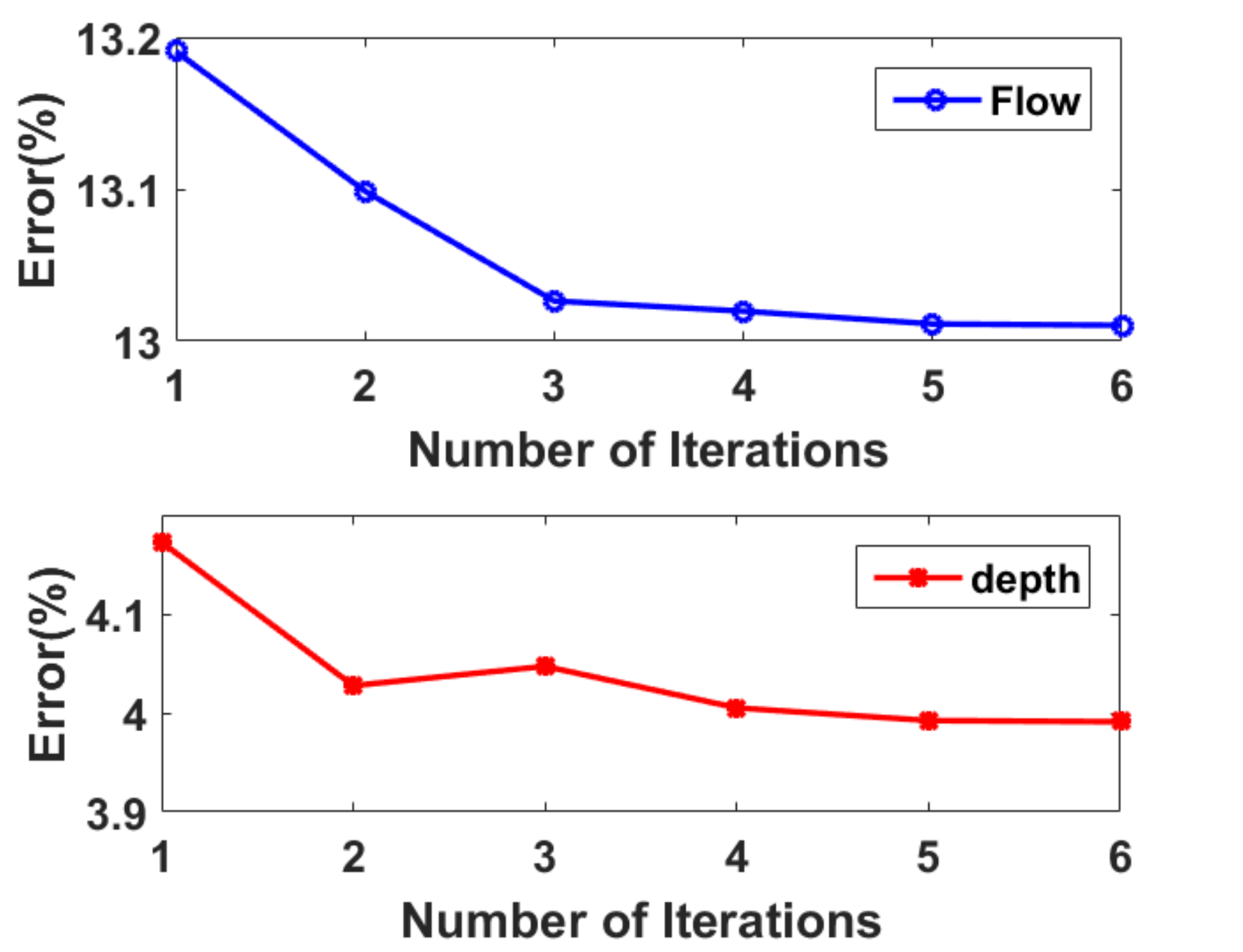}
\includegraphics[width=0.235\textwidth]{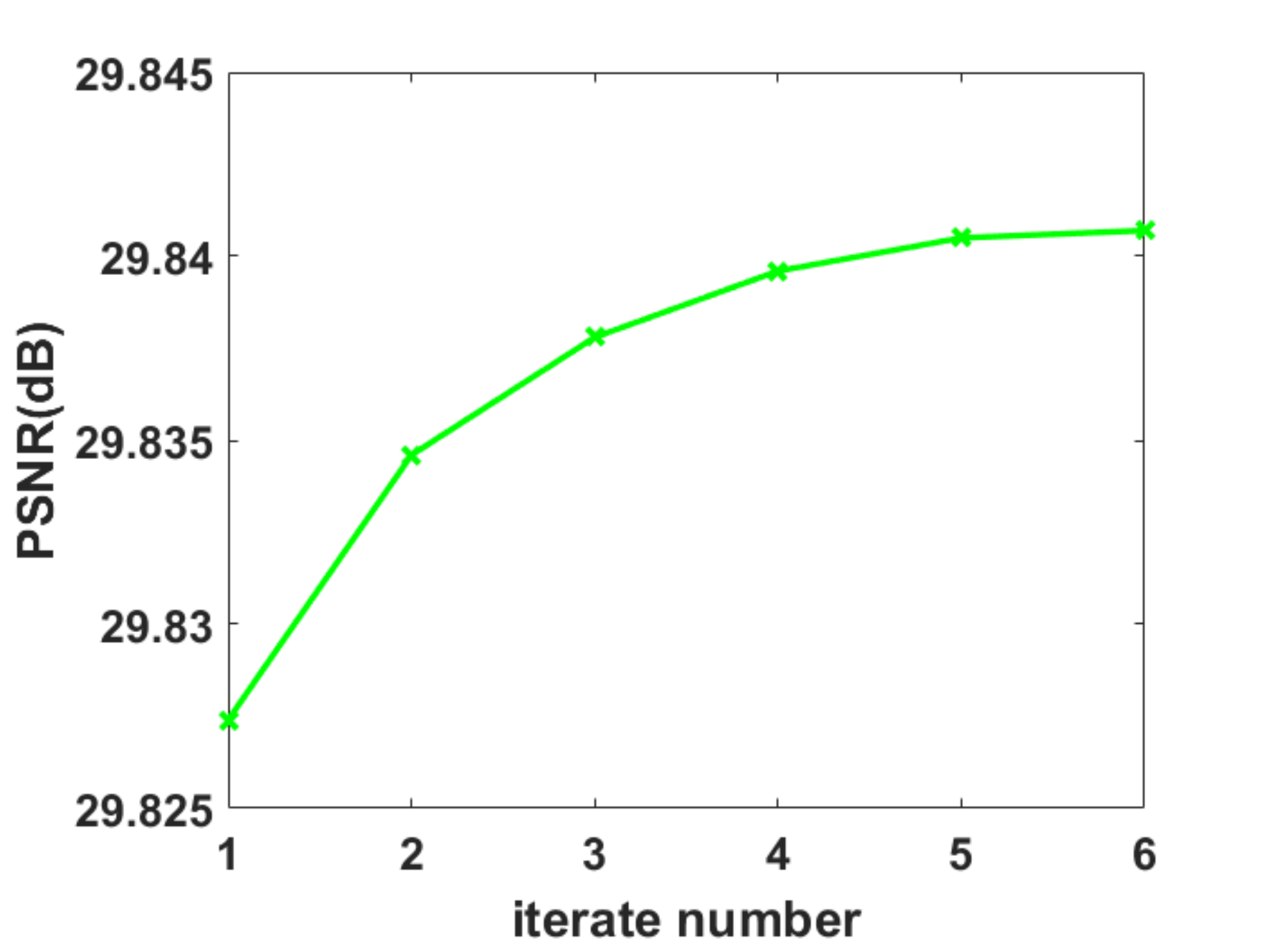}
\end{center}
\caption{Performance of our method on KITTI (flow error, depth error, PSNR) with the respect to the number of iterations.}
\label{fig:iter_fdpsnr}
\end{figure}

\noindent{\bf Results on KITTI.}
To the best of our knowledge, currently there is no realistic benchmark datasets that provide blurry images and corresponding ground-truth depth maps and the latent clean images. In this paper, we take advantage of the KITTI visual odometry dataset~\cite{geiger2013vision} to create a synthetic blurry image dataset on realistic scenery, where each sequence includes 6 images ($375 \times 1242$). Then we obtained the depth sequence with a down-sample factor $r = 4$. The blurry images are generated by using the piecewise linear kernel, where frame rate is set as $\tau = 0.23$ and the number of frame is $N = 20$. Therefore, the image blur is caused by both objects motion and camera motion with occlusion and shadow. We perform block-coordinate-descent on a subset of 30 randomly selected training images to obtain the optimal model parameters $\{w,c\}$ and $\{\alpha\}$ in cross-validation.

We evaluated results on the reference image and our method consistently outperforms all baselines as illustrated in Table~\ref{all_all}. We achieve the minimum bad pixel ratio of 3.91\% for depth completion and a PSNR of 29.83 for image deblurring. Fig.~\ref{fig:kitti_all} shows qualitative depth completion results and deblurring results of of our method and other competing methods on the sample sequences from the KITTI dataset.

Fig.~\ref{fig:iter_fdpsnr} shows the performance of our depth completion and image deblurring method with respect to the number of iterations, where the performance of both depth completion and image deblurring improves with the increase of iterations. While we use 6 iterations for all our experiments, the experiments indicate that 3 iterations are sufficient in most cases to reach an optimal performance for our formulation.

\noindent{\bf Results on TUM.}
In order to evaluate the performance of our method on real dataset, we also use the TUM RGB-D dataset~\cite{sturm2012benchmark} which included motion blur. The captured depth maps and color images are of size $640 \times 480$. Then we down-sample the obtained depth maps with rate $r = 16$ to simulate sparse depth maps. We evaluated our results on the reference image and achieve the minimum bad pixel ratio of 0.22\% for depth completion, consistently outperforms all baseline methods. Fig.~\ref{fig:bear_all} shows the visually completed depth map and deblurring results of our method comparing with other methods on sample sequences from the TUM dataset.
\begin{table}[h]\footnotesize
\centering
\caption{Quantitative evaluation on the KITTI dataset where the blur images are generated by averaging three consecutive frames.}
\label{kittiaver_psnr}
\begin{tabular}{c|c|c|c}
\hline
                                  & PSNR(dB)     & SSIM(\%)     & Depth Error(\%)\\ \hline
Yang  \etal \cite{yang2014color}  &   /          &   /          &  6.15 \\ \hline
D Ferstl \etal \cite{ferstl2013image}  &  /      &   /          &  3.22 \\ \hline
J Park \etal \cite{park2014high}   &  /          &   /          &  9.63 \\ \hline
Kim and Lee ~\cite{hyun2015generalized} & 23.21  & 0.781        &  /    \\ \hline
Sellent \etal \cite{sellent2016stereo}  & 23.31  & 0.764        &  /    \\ \hline
Ours                          & \bf23.89         & \bf0.786     &  \bf2.77     \\ \hline
\end{tabular}
\end{table}

\noindent{\bf Results on Another Blur Model.}
Even though the TUM dataset contains blurry images, they cannot be used for quantitative evaluation since no ground truth clean images are available. To perform such quantitative evaluation, synthetic images have been widely used ~\cite{hyun2015generalized,nah2016deep, gong2016motion}. We have evaluated our method under the spatial-variant blur generation model. Here we tested our method on another blur generation model (the blur image is simply an average of consecutive three frames). The results are shown in Table \ref{kittiaver_psnr}, where our method again achieves the best performance.

\section{Conclusion}
In this paper, we present a joint optimization framework to tackle the challenging task of depth map completion with the guidance of blurry color images, where depth completion and sequence images deblurring are solved in a coupled manner. Under our formulation, the motion cues from depth completion and blurry images could benefit each other, and produce superior results than conventional depth completion or deblurring methods. The performance of our method has been evaluated on both outdoor and indoor scenarios.

\vspace{-0.1cm}
\section*{Acknowledgement}
\vspace{-0.1cm}
This work was supported in part by China Scholarship Council (201506290130), Australian Research Council (ARC) grants (DP150104645, DE140100180), and Natural Science Foundation of China (61420106007, 61473230, 61135001), State Key Laboratory of Geo-information Engineering (NO.SKLGIE2015-M-3-4), Natural Science Foundation of Shaanxi (2017JM6027, 2017JQ6005) and Aviation fund of China (2014ZC53030). We thank all reviewers for their valuable comments.
\balance

{\small
\bibliographystyle{plain}
\bibliography{Deblur-Reference.bib}
}

\end{document}